\documentclass[journal]{IEEEtran}
\usepackage[pdftex]{graphicx}
\usepackage[caption=false,font=footnotesize]{subfig}
\usepackage{amsmath,bm}
\usepackage{amssymb}
\usepackage{booktabs}
\usepackage[all]{nowidow}
\usepackage{stfloats}
\usepackage{comment}
\usepackage{cite}
\usepackage{url}
\usepackage{makecell}
\usepackage{array}
\usepackage{multirow}
\usepackage{enumitem}
\usepackage[export]{adjustbox}
\usepackage[pagebackref=false,breaklinks=true,colorlinks,bookmarks=false]{hyperref}
\usepackage{xspace}
\usepackage[table,dvipsnames]{xcolor}
\usepackage{bbm}
\usepackage[super]{nth}
\usepackage{dirtytalk}
\usepackage{soul}
\usepackage[normalem]{ulem}

\xspaceaddexceptions{]\}}

\newcommand{\utilise}{\mbox{U-TILISE}\xspace}
\newcommand{\earthnet}{\mbox{EarthNet2021}\xspace}
\newcommand{\senXIImscrts}{\mbox{SEN12MS-CR-TS}\xspace}

\newcommand{\sentinelII}{\mbox{Sentinel-2}\xspace}
\newcommand{\ltae}{\mbox{L-TAE}\xspace}
\newcommand{\sits}{time series\xspace}
\newcommand{\Sits}{Time series\xspace}
\begin{document}

\title{\utilise: A Sequence-to-sequence Model for Cloud Removal in Optical Satellite Time Series}

\author{Corinne~Stucker,
        Vivien~Sainte~Fare~Garnot,
        and~Konrad~Schindler,~\IEEEmembership{Senior~Member,~IEEE}%
\thanks{Corinne Stucker and Konrad Schindler are with the Chair of Photogrammetry and Remote Sensing, ETH Z\"urich, 8093 Z\"urich, Switzerland (e-mail:
stuckerc@ethz.ch; schindler@ethz.ch).}%
\thanks{Vivien Sainte Fare Garnot is with the Institute for Computational Science, University of Z\"urich, 8057 Z\"urich, Switzerland (e-mail: vsaint@ics.uzh.ch).}
} %

\markboth{}%
{}

\maketitle
\begin{abstract}
Satellite image time series in the optical and infrared spectrum suffer from frequent data gaps due to cloud cover, cloud shadows, and temporary sensor outages. It has been a long-standing problem of remote sensing research how to best reconstruct the missing pixel values and obtain complete, cloud-free image sequences.
We approach that problem from the perspective of representation learning and develop \utilise, an efficient neural model that is able to implicitly capture spatio-temporal patterns of the spectral intensities, and that can therefore be trained to map a cloud-masked input sequence to a cloud-free output sequence. The model consists of a convolutional \emph{spatial encoder} that maps each individual frame of the input sequence to a latent encoding; an attention-based \emph{temporal encoder} that captures dependencies between those per-frame encodings and lets them exchange information along the time dimension; and a convolutional \emph{spatial decoder} that decodes the latent embeddings back into multi-spectral images. 
We experimentally evaluate the proposed model on \earthnet, a dataset of \sentinelII time series acquired all over Europe, and demonstrate its superior ability to reconstruct the missing pixels. Compared to a standard interpolation baseline, it increases the PSNR by 1.8$\,$dB at previously seen locations and by 1.3$\,$dB at unseen locations.
\end{abstract}


\IEEEpeerreviewmaketitle

\section{Introduction}
\IEEEPARstart{M}{odern} satellite images have made it possible to continuously and systematically monitor the Earth's surface.
Remotely sensed imagery, and products derived from it by image classification~\cite{russwurm2020self,garnot2020satellite,hong2021spectralformer},
segmentation~\cite{garnot2021panoptic,tarasiou2023vits}, or regression~\cite{potapov2021mapping,becker2023country,daudt2023snow}, have become an important data source for applications ranging from environmental monitoring~\cite{song2018global,wernick2021quantifying,lang2022high} to agricultural management~\cite{mulla2013twenty,hank2019spaceborne,segarra2020remote}. Moreover, multi-temporal satellite image sequences provide unprecedented opportunities to explore the temporal dynamics of natural processes, such as the evolution of land cover phenology~\cite{bolton2020continental}.
Among Earth observation missions that systematically revisit the same locations and capture such sequences, most acquire images in the optical and near-infrared part of the electromagnetic spectrum. On the one hand, optical images are particularly suitable for visual interpretation by humans. On the other hand, spectral signatures in that part of the spectrum contain information that makes it possible to distinguish between different land cover types and to quantify the health and vitality of vegetation. Notably, several indicators for vegetation density and productivity are based on (heuristic, non-linear) combinations of spectral intensities between 400$\,$nm and 1500$\,$nm. Such indices, including the widely used normalized difference vegetation index (NDVI)~\cite{rouse1974monitoring}, are a mainstay of current remote sensing practice~\cite{misra2020status,garioud2021recurrent}. 

Unfortunately, the effective availability of optical satellite images is considerably lower than the nominal revisit frequency of the satellites. Data gaps routinely occur, either due to occlusions or because no image is captured during an overpass for technical reasons, such as sensor maintenance or conflicting imaging requests. The primary cause for data gaps is the weather, i.e., clouds, haze, and cloud shadows that partially or fully obscure the observed scene. A study over 12~years of optical data acquired by the MODIS sensor concluded that, on average, clouds occlude 67\% of the Earth's surface and 55\% of the land surface at any point in time~\cite{king2013spatial}. The large proportion of data gaps, which moreover are irregularly distributed, calls for measures to ensure the usability of monitoring systems in the presence of frequent clouds.

A first, rudimentary approach adopted by several satellite processing pipelines is to discard images affected by clouds before further analysis. Processing solely cloud-free image observations may make data handling and visual inspection more convenient, but it also discards a lot of data that may still be usable, since often only a moderate fraction of an image or time series is affected by clouds. Moreover, \cite{gawlikowski2022explaining} has shown that learning-based image classifiers trained on curated, cloud-free training datasets do not perform all that well when applied to images with even a small amount of clouds, let alone to images with moderate or severe cloud cover. More recent pipelines operate on all available input images, regardless of their degree of cloud cover, and learn to ignore uninformative pixels at an algorithmic level, for instance, via data-driven attention mechanisms~\cite{russwurm2020self}.
An alternative strategy is to \emph{remove} clouds and cloud shadows in advance, such that the subsequent processing is no longer affected by them, rather than (explicitly or implicitly) ignore them during image analysis. An advantage of such a two-step approach is that multiple image analysis pipelines tailored to different tasks can all use the same cloud-free images. These pipelines can then be more efficient and often also more robust because they need not each devote a significant share of their model capacity to the detection and handling of data gaps~\cite{gu2022explicit}.

In the last few years, encoder-decoder style neural networks have become the prevalent approach to recover missing data in optical satellite images. Several methods have tried to directly learn the mapping from a cloudy to a cloudless RGB image~\cite{enomoto2017filmy,grohnfeldt2018conditional,singh2018cloud}, others even try to translate Synthetic Aperture Radar (SAR) images to multi-spectral optical images~\cite{bermudez2018sar,bermudez2019synthesis,wang2019sar,fuentes2019sar,zhang2020feature}. A recent trend has been to combine the two inputs and perform SAR-optical data fusion~\cite{meraner2020cloud,ebel2022sen12ms,xu2022glf}. SAR systems employ active illumination with much longer wavelengths, which are unaffected by clouds and shadows and may provide complementary information to guide the reconstruction of occluded content. However, one must bridge a considerable domain gap to infer optical reflectance values from SAR amplitudes.
Both are likely to change at land cover boundaries, so SAR may help to restore image gradients, but it can hardly be expected to offer much information about the actual spectral values, colors, and texture details. In contrast, sequences of optical images depicting the same location%
\footnote{Throughout this article, the shorthand \say{time series} denotes a temporally ordered, co-registered sequence of images showing the same location.} %
exhibit strong correlations along the temporal dimension. These spatio-temporal patterns provide evidence about the spatial structure \emph{and} the spectral properties. Importantly, this holds not only for static or gradually changing land cover but also includes more complex temporal dynamics, like seasonal variations. The idea to leverage multi-temporal images for cloud removal is rather obvious; but most existing approaches~\cite{sarukkai2020cloud,zhang2020thick,oehmcke2020creating,chen2020thick,ebel2022sen12ms} collapse the multi-temporal, cloudy input into a single gap-free image, often additionally relying on spatially and temporally co-registered SAR observations to guide the reconstruction~\cite{ebel2022sen12ms}.

Here, we focus on full \emph{sequence-to-sequence} translation: our goal is to convert the cloudy input time series into a gap-free product with the same time steps, but containing a clean, cloud-free image at every frame (no matter whether the original frame was cloud-free, cloudy, or entirely missing).
To that end, we introduce \utilise,\footnote{\underline{U}-Net \underline{T}emporal \underline{I}mputation \underline{L}ightweight \underline{I}mage \underline{S}equence \underline{E}ncoder.} a neural image sequence model that captures spatio-temporal relationships between the spectral intensities in an image time series, and is, therefore, able to impute missing pixels. 
\utilise operates in three dimensions, with 2D~convolutions to encode multi-scale local relationships in space and 1D~(self-)attention to encode non-local relations in time.
By design, its output has the same spatial and temporal extent as the input, such that it jointly reconstructs complete, gap-free time series.
The model supports additional, auxiliary input channels and is therefore, in principle, able to use SAR amplitudes, too. But empirically this brings only a tiny improvement (see appendix).

We experimentally test \utilise on the \earthnet dataset~\cite{requena2021earthnet2021}, which contains thousands of \mbox{30-frame}, \mbox{4-channel} (R, G, B, NIR) time series of \sentinelII data. Compared to standard interpolation between the temporally nearest unoccluded observations, our model improves the peak signal-to-noise ratio (PSNR) of the reconstructed spectral values by 1.8$\,$dB at previously observed locations, and by 1.3$\,$dB at unseen locations.

The remainder of this article is organized as follows: first, we provide an overview of cloud removal methods in Section~\ref{sec:related_work}, with a focus on learned (mono-temporal as well as sequence-based) approaches. In Section~\ref{sec:methodology}, we introduce the \utilise model, set out its components (Sections~\ref{sec:architecture} and \ref{sec:pe}), and describe the associated training and inference procedures (Section~\ref{sec:training_inference}). Next, we explain the data (Section~\ref{sec:data}) and the experimental setup (Section~\ref{sec:experimental_setup}) used in our evaluation. Section~\ref{sec:results} reports and discusses experimental results, followed by a conclusion (Section~\ref{sec:conclusion}). Complementary experiments with additional SAR observations on top of optical \sits are given in the appendix.

\section{Related Work}
\label{sec:related_work}
Reconstructing missing pixels in remotely sensed imagery has been a long-standing research problem. Early efforts toward thin cloud and haze removal build on
physical relations~\cite{long2013single,xu2015thin,lv2016empirical}  
or signal processing considerations~\cite{shen2014effective,xu2019thin} to describe the process of light transmission and interaction with clouds. Methods designed to recover image content occluded by thick clouds have been based on tensor factorization~\cite{ng2017adaptive,li2019cloud,lin2022robust}, on mosaicking of multi-temporal images~\cite{Helmer2005Cloud,tseng2008automatic,lin2012cloud,lin2013patch} or they adopt statistical image processing methods originally developed for single-image inpainting~\cite{shen2008map,maalouf2009bandelet, meng2017sparse}. In the following, we concentrate on data-driven, learning-based methods for cloud removal. For completeness, we note that video inpainting methods like \cite{zeng2020learning} share conceptual similarities with satellite time series imputation, but they are beyond the scope of the present literature review.

\subsection{Mono-temporal cloud removal}
A natural formulation of cloud removal is as an image-to-image translation task, where the mapping from the cloudy input to the cloud-free output is learned in a data-driven manner. For instance,~\cite{enomoto2017filmy} employ a conditional generative adversarial network (cGAN)~\cite{mirza2014conditional} to map from a cloudy to a cloudless RGB satellite image. The mapping is conditioned on the NIR channel of the input, arguing that near-infrared wavelengths partially penetrate clouds and may thus capture information about the observed scene that is unavailable in the visible spectrum. In~\cite{grohnfeldt2018conditional}, the NIR channel is replaced with conditioning on a SAR image, as clouds are completely transparent at radar wavelengths. The works of \cite{enomoto2017filmy,grohnfeldt2018conditional} both do not go beyond a proof-of-concept; the underlying neural networks are trained and evaluated exclusively on synthetically generated images with clouds simulated by Perlin noise~\cite{perlin2002improving}. It has since been shown that such simulations generalize poorly to real cloudy images~\cite{ebel2021multisensor}. To side-step the need for training examples where cloudy and cloud-free images are in exact, pixel-wise correspondence, \cite{singh2018cloud,ebel2021multisensor} rely on a cycle-consistency loss~\cite{zhu2017unpaired}. In this way, the networks can be trained directly on images with real data gaps, eliminating the potential domain gap between training and test data. While the method in \cite{singh2018cloud} is limited to thin clouds, \cite{ebel2021multisensor} do not impose any restrictions on the maximum permissible cloud coverage or density. Furthermore, \cite{ebel2021multisensor} combine explicit modeling of cloud densities with a residual learning strategy to better preserve the pixel values in cloud-free image regions. The methods mentioned so far have in common that they are limited to images with three optical bands. To address that limitation, several SAR-to-optical image translation approaches~\cite{bermudez2018sar,bermudez2019synthesis,wang2019sar,fuentes2019sar,zhang2020feature} learn the mapping from a SAR image to the full stack of multi-spectral bands, often also using cGANs.

Recent advances in learned cloud removal tend to rely on image fusion, i.e., they synergistically use the cloudy optical and a cloud-free SAR image to impute the missing pixels in the former. In \cite{meraner2020cloud}, a \sentinelII image and a temporally close SAR image of the same scene are stacked together along the channel dimension and fed into a neural network that regresses a residual reflectance value at every pixel. Those per-pixel corrections are then added to the input to remove the missing data. \cite{gao2020cloud} combine SAR-to-optical translation and SAR-optical data fusion in a cascaded fashion. First, a GAN is trained to map the SAR input to an optical image. That synthetic image is stacked with the original SAR data and the cloudy optical input and fed into a second GAN, trained to map the multi-modal input to a cloud-free optical image.

The recent \cite{xu2022glf} found that stacking optical and SAR observations into a multi-modal image and processing them together does not optimally exploit the two inputs, as feature extraction from SAR is aggravated by speckle noise. Instead, the authors propose separate embedding branches per modality, together with an attention-based mechanism that gradually and selectively fuses features from the two branches.

SAR-optical data fusion approaches have demonstrated that complementary information in the form of SAR observations can help to compensate for missing data in optical images. Still, fusing optical and SAR data remains challenging due to the large domain gap between the two modalities. One must also keep in mind that SAR can hardly contribute to restoring actual spectral reflectance information, like different hues or fine-grained textures. Its role is to add spatial context, such as land-cover boundaries, which appear as gradients also in the SAR amplitude. 
An alternative approach is to inject contextual information from other optical sensors, as, for instance, in~\cite{liu2023thick}. Clearly, this will greatly reduce the domain gap, but on the other hand, there is no guarantee that a temporal close and largely cloud-free image can be found.

\subsection{Sequence-based cloud removal}
Sequence-to-point methods~\cite{sarukkai2020cloud,zhang2020thick,oehmcke2020creating,chen2020thick,yang2022deep,sebastianelli2022plfm,ebel2022sen12ms} consume a multi-spectral time series with data gaps and output a single, gap-free image. In most cases, that output does not have a well-defined time stamp but rather is seen as representative of the entire time period between the start and end dates of the time series.
Even if the output is associated with a specific time, e.g., the middle frame, the method would have to be run iteratively to reconstruct an entire time series.
Typically, the input time series only have three to five images. Optionally, the reconstruction can additionally be guided by a single SAR image~\cite{sebastianelli2022plfm} or by a SAR time series (approximately) aligned frame-to-frame with the input~\cite{ebel2022sen12ms}.
Some of these sequence-to-point approaches impose tight restrictions on the maximum cloud coverage per image, e.g., \cite{sarukkai2020cloud} require at most \mbox{10--30\%} cloud cover, and \cite{yang2022deep} require the first and last of three frames to be cloud-free (0\% cover).
Furthermore, these methods often assume short temporal intervals with minimal land cover changes over time since they accumulate spectral information along the temporal dimension to create the output image. That assumption rules out systematic land cover dynamics, in particular seasonal cycles of vegetation and agriculture. 

To our knowledge, \cite{ebel2021internal,peng2022reconstruction,zhao2023seeing} are the only published sequence-to-sequence models, i.e., they output a cloud-free, multi-spectral image for every frame of the input time series. In \cite{peng2022reconstruction}, the sequence model is parameterized as a recurrent neural network with a two-layer GRU~\cite{cho2014learning} architecture and used to learn the mapping from a SAR time series to a time series of Landsat images, only for pixels belonging to a specific land cover class, namely, rice fields. This yields reconstructions of limited quality (PSNR \textless$\,$28$\,$dB), possibly due to the well-known difficulties of learning multi-layer recurrent models~\cite{turkoglu2021gating}.
A rather different approach is taken by~\cite{ebel2021internal}, who adopt a recent video inpainting technique~\cite{zhang2019internal} that extends the deep image prior (DIP)~\cite{ulyanov2018deep} to videos. The initial input is a SAR time series (as opposed to random noise in the original DIP) such that the network effectively performs a mapping from SAR to optical time series.
We point out that while the DIP employs a neural network parameterization, it is not a learned model. The convolutional network structure, which favors lower amplitudes for high-frequency signals, serves as a hard-wired low-level prior of optical image statistics. It does not store any a priori information extracted from training data. Instead, the network weights are optimized individually for every input sequence at inference time.
The method most related to our proposed \utilise model is \cite{zhao2023seeing}, an adversarial approach that internally splits the computation into a first, coarse round of imputation and a subsequent refinement network. The coarse imputation model is a 3D~spatio-temporal encoder-decoder architecture with separate backbones for the optical and SAR inputs, followed by a transformer-style attention mechanism in the bottleneck to fuse the latent embeddings of the two encoder branches. The model does not implicitly learn to ignore cloudy observations; instead, the contribution of cloudy input pixels is suppressed by explicitly modulating the learned attention masks according to the given cloud masks before applying them to the latent embeddings coming out of the optical encoder branch.

\section{Method}
\label{sec:methodology}

\begin{figure*}[!ht]
\centering
\includegraphics[width=0.8\textwidth]{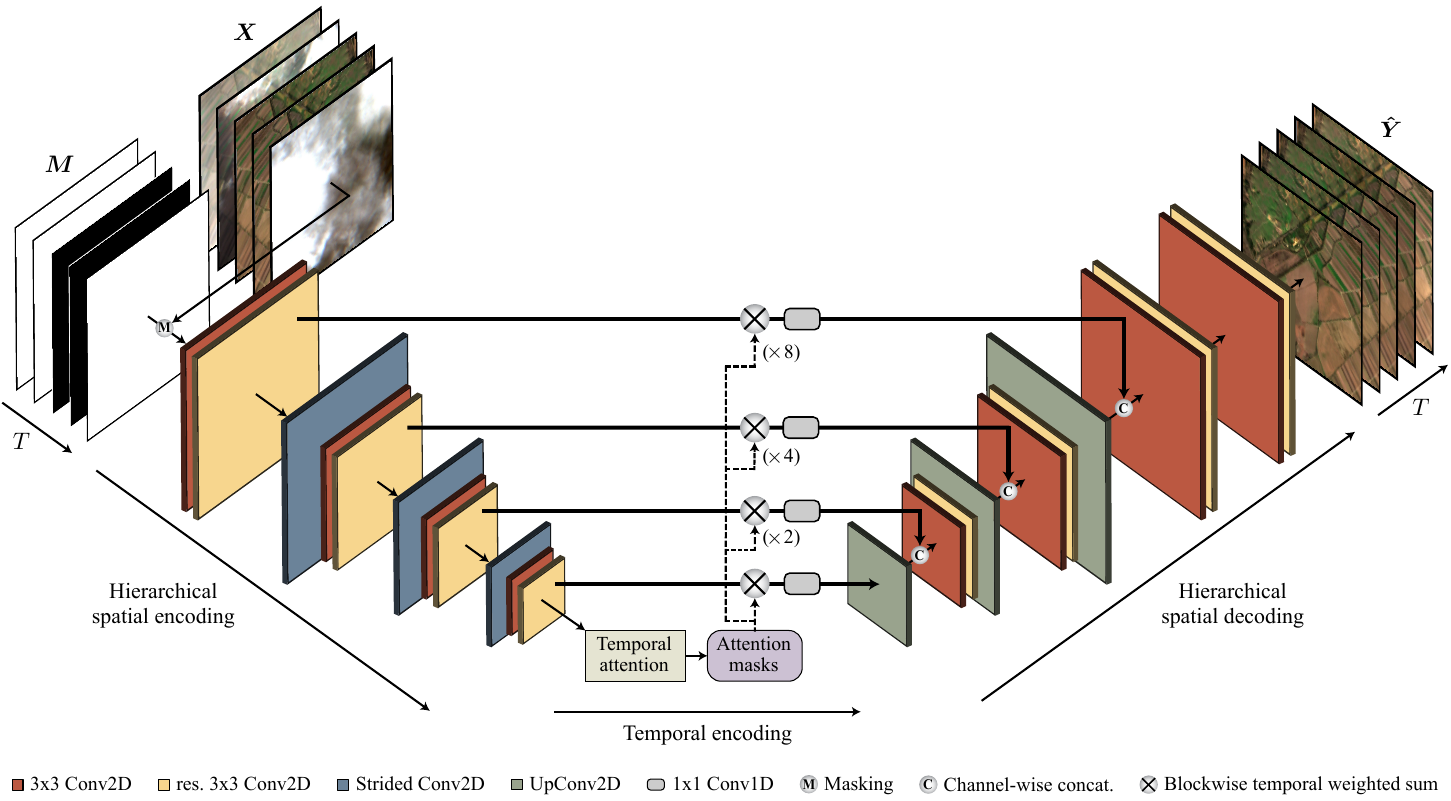}
\caption{Overview of the proposed model: \utilise is a neural sequence-to-sequence model that takes as input a multi-spectral satellite image time series in which missing reflectance values are masked and outputs a complete and gap-free \sits with the same dimensions. \utilise employs a convolutional U-Net architecture~\cite{ronneberger2015u} over the spatial and spectral dimensions and a transformer-style self-attention mechanism along the temporal dimension. The attention masks operate on the (spatial) bottleneck between the encoder and decoder parts of the U-Net as well as on the skip connections (after suitable upsampling).}
\label{fig:overview_utilise}
\end{figure*}

To impute missing image content, we design \utilise, a learnable image sequence model in the form of a neural network. Once trained, the weights of that network encode a prior over spatio-temporal patterns of multi-spectral reflectance. When a \sits with data gaps is fed into the network, the prior fills in the missing values to obtain a complete, gap-free \sits.
In contrast to existing cloud removal pipelines, our approach does not rely on auxiliary SAR observations to guide the imputation. Instead, we exploit spatial and temporal patterns within the multi-spectral \sits itself to reconstruct the spatio-temporal evolution of the depicted land cover. Furthermore, our model jointly reconstructs all images in a given time series, as opposed to pipelines that reconstruct a single frame, considered representative of the entire sequence.

\subsection{Problem formulation}
Let ${\bm{X} \in \mathbbm{R}^{T \times C \times H \times W}}$ denote a multi-spectral \sits, represented as a 4D~tensor with $T$ the temporal length, $C$ the number of spectral bands, and ${H \times W}$ the spatial extent. Our goal is to regress a reflectance for every spatio-temporal location, so as to obtain a complete, gap-free multi-spectral \sits ${\bm{\hat{Y}} \in \mathbbm{R}^{T \times C \times H \times W}}$. Our model assumes that the spatio-temporal locations to be imputed are marked by a binary mask ${\bm{M} \in \mathbbm{R}^{T \times 1 \times H \times W}}$, where the value~1 denotes pixels with a valid observation and~0 denotes missing data values. Note that we do not impose any assumptions or requirements on the mask $\bm{M}$: it may denote any type of data gaps, including clouds and cloud shadows, but also frames that are entirely missing, for instance, due to sensor maintenance.

\subsection{Overview}
Fig.~\ref{fig:overview_utilise} gives a graphical overview of our method. At its core is \utilise, a learned sequence-to-sequence model that captures spatio-temporal relationships between the spectral intensities, and thereby is able to reconstruct realistic, complete, and gap-free multi-spectral \sits. For efficiency, the mask ${\bm{M} \in \mathbbm{R}^{T \times 1 \times H \times W}}$ of missing pixels is not separately fed into the model but imprinted directly on the multi-spectral input ${\bm{X} \in \mathbbm{R}^{T \times C \times H \times W}}$ by setting all masked pixels to the maximum intensity~1. \utilise consists of three components. First, a shared multi-scale convolutional encoder transforms every image of the masked sequence into a latent embedding. Next, an attention-based temporal encoder combines the per-frame embeddings across time to impute missing values in the latent sequence representation. Last, a shared convolutional decoder reconstructs every image from its latent embedding to obtain a gap-free \sits with the spatial, spectral, and temporal dimensions of the input.

\subsection{3D spatio-temporal sequence-to-sequence model}
\label{sec:architecture}
\utilise builds on recent advances in learned \sits processing. Its architecture is inspired by \mbox{U-TAE}~\cite{garnot2021panoptic}, a model originally developed for crop mapping, which maps a \sits to a (mono-temporal) panoptic segmentation. In a nutshell, U-TAE combines convolutions for multi-scale spatio-spectral encoding with a lightweight non-local temporal attention mechanism~\cite{garnot2020satellite}. Intuitively, the latter learns to focus on the most salient observations within a given \sits. By design, \mbox{U-TAE} collapses the input along the temporal dimension to produce a mono-temporal output. We take inspiration from the design of modern transformer models~\cite{devlin2018bert,xiong2020layer} and extend the architecture to a full 3D~spatio-temporal sequence-to-sequence model that preserves the temporal dimension.

\utilise consists of \textit{(i)} symmetric multi-scale spatio-spectral encoding and decoding modules in the style of \mbox{U-Net}~\cite{ronneberger2015u} and \textit{(ii)} a lightweight temporal encoding module based on multi-head self-attention~\cite{vaswani2017attention}, see Fig.~\ref{fig:overview_utilise}. We now describe each of these components in more detail.

\subsubsection{Spatial encoder}
The spatial encoder gradually transforms the masked \sits of size $(T \times C \times H \times W)$ into a multi-scale latent embedding via a sequence of convolutional blocks. Each block comprises a ${3 \times 3}$ convolutional layer with stride~1 and $d$ filter channels, followed by a rectified linear unit (ReLU) as non-linear activation function and a residual  ${3 \times 3}$ convolution with stride~1, $d^{\prime}$ filter channels, and ReLU activation. Between the convolutional blocks, a strided convolution equipped with ReLU activation decreases the spatial resolution of the intermediate embeddings by a factor of 2. After encoding all images in the time series (individually and in parallel), we temporally stack their latent representations to produce a multi-temporal sequence embedding with dimensions $(T \times D \times H/8 \times W/8)$, with $D$ the channel depth of that embedding.

\subsubsection{Temporal encoder}
The temporal encoder operates individually on the spatial locations (low-resolution \say{pixels}) of the latent embedding. For each such pixel, it captures the pairwise dependencies between the values in all pairs of different frames and uses them to fill in missing information. The temporal encoder is based on the Lightweight Temporal Attention Encoder (\ltae) of \cite{garnot2020satellite}, which, in turn, is a simplified version of the multi-head self-attention mechanism of the transformer architecture~\cite{vaswani2017attention}. Unlike  \cite{garnot2020satellite}, we employ data-driven queries to preserve the temporal dimension of the input. Moreover, we use residual skip connections as in the original transformer model \cite{vaswani2017attention}. We retain the computational simplifications of \cite{garnot2020satellite} and use a channel grouping, where the $G$ attention heads process mutually exclusive subsets of ${D/G}$ channels of the embedding. The learned attention scores are directly applied to the embedding vectors that come from the encoder (without first modulating them with a fully-connected layer). Following recent findings about neural sequence-to-sequence models, we prefer pre-normalization with the \emph{groupnorm} scheme~\cite{wu2018group}. Furthermore, we employ GELU activations~\cite{hendrycks2016gaussian} rather than the classical ReLU activations in the multi-layer perceptron (MLP) of the attention block.

\subsubsection{Spatial decoder}
After the latent representation has been passed through the attention module, the spatial decoder progressively restores multi-spectral images from the individual per-image embeddings. These images have the same spectral and spatial resolution as the input to the network but no more missing values. The structure of the spatial decoding blocks is the same as for the spatial encoding blocks, except that fractionally strided, transposed convolutions with stride~$\frac{1}{2}$ replace the strided convolutions. Once the native spatial resolution of the input has been reached, a final convolutional block maps the latent embedding to the spectral space. The final layer uses sigmoid activations instead of ReLU, so as to regress reflectances in the range $[0,1]$. Finally, the reconstructed frames are stacked along the temporal dimension to recover the complete, gap-filled \sits.

\subsubsection{Skip connections}
Skip connections from encoder to decoder levels of equal spatial resolution are a key component of the \mbox{U-Net}~\cite{ronneberger2015u} architecture to propagate high-frequency details and localization information that is lost during spatial downsampling operations. We adopt the same strategy as in \cite{garnot2021panoptic} and temporally weight the information transferred between corresponding layers of the spatial encoder and decoder. The attention masks learned by the temporal encoder serve as weights, which we spatially upsample to the adequate spatial resolution using bilinear interpolation. The temporally weighted output of the encoding layers is processed with a shared ${1 \times 1}$ convolutional layer followed by ReLU activation before channel-wise concatenation with the output of the corresponding decoding layers for further processing.

\subsection{Sinusoidal positional encoding}
\label{sec:pe}
By itself, the self-attention mechanism is agnostic to the sequence order. To provide positional information, we follow the standard procedure for transformers~\cite{vaswani2017attention} and add a positional encoding (PE) to the input of the temporal encoder before applying self-attention:
\begin{equation}
    \text{PE}\left(t, k\right) = \sin\left (\text{day}(t) / \tau^{\frac{2k}{D}} + \frac{\pi}{2} \bmod{\left(k, 2\right)} \right).
\end{equation}

$\text{PE}\left(t, k\right)$ consists of fixed sinusoidal functions with predefined wavelengths and describes the position of the $t^\text{th}$~observation in the sequence, with $D$ the channel depth of the embedding and $k$ the coordinates of the positional encoding. We set $\tau= 1\,000$, as in \cite{garnot2020satellite}. Contrary to \cite{vaswani2017attention}, we do not directly encode the ordinal position $t$ in the sequence. Instead, we encode the observation date $\text{day}(t)$, expressed as the number of days since the $1^\text{st}$ of January of the respective calendar year. This strategy has proved beneficial for learned \sits processing~\cite{garnot2020satellite,metzger2021crop}, since it preserves information about seasonal patterns (e.g., lighting conditions or phenology of the vegetation) and accounts for irregular temporal sampling.

\subsection{Training and inference}
\label{sec:training_inference}
It is not possible to quantitatively assess the performance of \sits imputation for real data gaps due to the lack of ground truth reflectances. Therefore, we train and evaluate \utilise with simulated data (generated by masking cloud-free frames with real cloud masks taken from other sequences, cf.\ Section~\ref{sec:simulation}) and examine its capability to generalize to sequences with actual data gaps due to real clouds. We train \utilise in a supervised manner by minimizing the pixel-wise absolute differences between the imputed \sits~$\bm{\hat{Y}}$ and the corresponding ground truth \sits~$\bm{Y}$:
\begin{equation}
    \mathcal{L}_1 = \frac{1}{NCHW}\sum_{i=c=x=y=1}^{N, C, W, H} \frac{1}{T^i} \sum_{t=1}^{T^i} \mid \hat{Y}_{tcxy}^i - Y_{tcxy}^i \mid,
\end{equation}
where $N$ denotes the number of training sequences and $T^i$ the length of the $i^\text{th}$~sequence.

We train \utilise for a fixed temporal length of ${T=10}$, i.e., the input is a \sits that comprises at most 10~images. Shorter sequences are padded with no-data frames. During training, longer sequences are randomly cropped if ${T^i > T}$. At test time, we retain the full time series and process it in one shot if ${T^i \le T}$, or in sliding window fashion if ${T^i > T}$.

\section{Data}
\label{sec:data}
We evaluate our method on \earthnet, a large, publicly available dataset of \sentinelII \sits. Note, however, that our method is sensor-agnostic and can adapt to the properties of different multi-spectral imaging sensors, given suitable training data.

\subsection{\earthnet dataset}
\label{sec:earthnet}
The \earthnet~\cite{requena2021earthnet2021} dataset was originally designed for satellite image forecasting, conditioned on future meteorological variables. It includes more than $32\,000$~\sentinelII \sits collected over Central and Western Europe from November~2016 to May~2020. Each time series consists of 30~images with Level-1C top-of-atmosphere (TOA) reflectances. The images are acquired in a regular temporal interval of five days, where acquisition dates without an observation are encoded as images of NaN values. Every image is composed of the four spectral bands B2~(blue), B3~(green), B4~(red), and B8~(near-infrared) and covers a spatial extent of 128$\times$128$\,$pixels (2.56$\times$2.56$\,$km in scene space), resampled to the resolution of 20$\,$m. For every observation, the dataset further includes a pixel-wise cloud probability map\footnote{Pixel-wise cloud probabilities are only available for the training data.} obtained via the \texttt{S2Cloudless} algorithm~\cite{baetens2019validation} and a binary cloud and cloud shadow mask based on heuristic rules similar to~\cite{meraner2020cloud}.

In our experiments, we reserve $\approx\,$20\% of the training sequences for validation, where training and validation tiles are mutually exclusive. For testing, we use the \textit{iid} and \textit{ood}~test splits. \Sits in the \textit{iid}~test split stem from the same \sentinelII tiles as the training data and the ones in the \textit{ood}~test split from previously unseen locations.

\subsection{Preprocessing}
\label{sec:cloud_filtering}
We adopt the preprocessing protocol of prior works~\cite{requena2021earthnet2021} and value-clip the optical images to the range ${[0, 10\,000]}$, followed by normalization to the unit range of ${[0, 1]}$.

To train a system for cloud removal that regresses \sits rather than a single image, we found experimentally that pixel-wise supervision for every spatio-temporal location is crucial for learning seasonal changes and land surface dynamics over time. Since obtaining such ground truth for \sits with real data gaps is impossible, we resort to cloud-free \sits and introduce synthetically generated data gaps during training and evaluation, as described in Section~\ref{sec:simulation}. Starting from a \sits with real data gaps, we first identify all images with partially occluded pixels or images that are occluded/missing entirely by applying a threshold to the cloud probability maps (if available) or the binary cloud masks. We choose the threshold in a conservative manner to minimize the number of undetected data gaps. We then remove all images with data gaps to produce cloud-free \sits that exhibit a valid\footnote{A few cloudy images may remain in the cloud-filtered \sits due to inaccuracies of the preceding cloud detector.} observation for every spatio-temporal location. Second, we discard \sits with less than five remaining images, as we deem such sequences too short for learning spatio-temporal patterns.
See the appendix for a summary of the number of \sits, their temporal lengths, and the temporal resolution before and after filtering images with data gaps.

\subsection{Simulation of data gaps}
\label{sec:simulation}
Realistically simulating cloud cover in satellite images is notoriously difficult. Synthetic images generated with existing physics-inspired simulation methods, like the well-known Perlin noise model~\cite{perlin2002improving}, do not match the radiometry of real data well enough: it has been shown that cloud removal methods trained with such synthetic images do not generalize well to images containing actual clouds \cite{ebel2021multisensor}. To create \sits with artificial data gaps for training and evaluation, we thus refrain from rendering synthetic clouds. Instead, we adopt a strategy commonly employed for image inpainting~\cite{yu2018generative,liu2018image,yu2019free} and completely mask out invalid pixels by setting them to the maximum intensity value~1. In this way, one only has to realistically simulate binary cloud masks, which is straightforward: all one needs to do is randomly sample real cloud masks from other acquisition times and/or locations within the same \sentinelII tile and apply them to a gap-free \sits. With that strategy, we obtain \sits with data gaps of realistic shapes and sizes and with known ground truth reflectances at all masked pixels in a controlled, fully automatic manner.

\section{Experiments}
\label{sec:experimental_setup}
\subsection{Setup}
Sequence-to-point methods typically use a temporally close cloud-free image of the same location to quantify the fidelity of the synthesized output. \cite{ebel2022sen12ms}, one of the few existing sequence-to-sequence approaches, regresses a time series of multi-spectral \sentinelII images but restricts the evaluation to a single time step, namely, the one with the lowest cloud cover. Such single-frame evaluation protocols are, in our view, problematic. On the one hand, they do not actually measure the quality of the regressed \sits, as the temporal aspect is completely ignored. On the other hand, metrics computed only from the least cloudy image will likely be too optimistic, since the imputation task becomes more challenging with increasing occlusions. 

We argue that the evaluation should take into account all images with missing pixels, irrespective of the degree of cloud cover. As ground truth reflectances are unavailable for \sits with real data gaps, we quantitatively evaluate \utilise on cloud-free sequences with synthetically added data gaps. Additionally, we apply the learned model (without further fine-tuning) to \sits featuring real data gaps to qualitatively assess performance in the true application setting. 

We follow the procedure described in Sections~\ref{sec:cloud_filtering} and \ref{sec:simulation} to generate \sits with synthetic data gaps. Unless stated otherwise, we randomly trim the cloud-filtered time series to a maximum length of ${T=10}$ images during training. At test time, we process the full-length sequences in sliding window fashion if their length exceeds~$T$. To simulate data gaps, we randomly superimpose at most 50\% of the images per time series with cloud (and cloud shadow) masks drawn randomly from the dataset, with a minimum of one masked image per sequence. When processing sequences with real data gaps, the actual cloud masks are used to mark missing pixels.

\subsection{Implementation details}
\utilise is implemented in PyTorch~\cite{paszke2019pytorch}. For training, we use an NVIDIA GeForce RTX 2080 Ti GPU for \sits with four spectral bands and an NVIDIA TITAN RTX GPU for \sits with 13 spectral bands. %
Source code and pretrained models are available at \url{https://github.com/prs-eth/U-TILISE}.

In our experiments, we use 64~filter channels for the spatial encoding and decoding convolutional layers, except at the lowest spatial resolution, where we use 128~filter channels. Accordingly, the temporal encoder has a latent feature dimension of 128. Temporal self-attention employs ${G=4}$ heads, with a dimension of four for the data-driven keys and queries. To augment the training data, we randomly rotate all images in a sequence by \mbox{$\alpha\in\{0^\circ,90^\circ,180^\circ,270^\circ\}$} and randomly flip them along the $x$ and $y$ axes.

We train with the Adam optimizer~\cite{adam} with hyper-parameters \{\mbox{$\beta_1$=0.9}, \mbox{$\beta_2$=0.999}\}, a batch size of three, and no weight decay. During the first 250~epochs, the initial base learning rate of \mbox{$2 \cdot 10^{-4}$} is halved every 50~epochs. Training is stopped once the \mbox{$\mathcal{L}_1$ loss} (cf.\ Section~\ref{sec:training_inference}) on the validation set has converged. In our experiments, this took about 1$\,$000~epochs, or 20~days of training on a single GPU. The computational cost for applying the trained model is low: the forward pass for a time series consisting of (at most) 10~images takes $\approx\,$0.02~seconds. Longer sequences are processed in sliding window fashion, which on average takes 0.13~seconds for a 30-frame sequence.

\subsection{Evaluation metrics}
We adopt a suite of metrics commonly used to evaluate cloud removal and inpainting methods: mean absolute error (MAE), root mean square error (RMSE), spectral angle mapper (SAM)~\cite{kruse1993spectral}, peak signal-to-noise ratio (PSNR), and structural similarity index (SSIM)~\cite{wang2004image}:
\begin{align}
    \text{MAE} &= \frac{1}{C\left|\Omega^i\right|}\sum_{c=1}^{C}\sum_{(t,x,y)\in \Omega^i} \mid \hat{Y}_{tcxy}^i - Y_{tcxy}^i \mid, \\
    \text{RMSE} &= \sqrt{\frac{1}{C\left|\Omega^i\right|}\sum_{c=1}^{C}\sum_{(t,x,y)\in \Omega^i}\left(\hat{Y}_{tcxy}^i - Y_{tcxy}^i\right)^2},  \\
    \text{SAM} &= \frac{1}{\left|\Omega^i\right|}\sum_{(t,x,y)\in \Omega^i}\! \arccos\left(\frac{\bm{\hat{Y}}_{txy}^i \cdot \bm{Y}_{txy}^i}{\| \bm{\hat{Y}}_{txy}^i\| \cdot \|\bm{Y}_{txy}^i\|}\right), \\
    \text{PSNR} &= 20 \cdot \log_{10}\left(\frac{1}{\text{RMSE}}\right),  \\
    \text{SSIM} &= \frac{1}{\left|T_m^i\right|}\sum_{t \in T_m^i} \frac{\left(2\mu_{\bm{\hat{Y}}_t^i}\mu_{\bm{Y}_t^i} \!+\! \epsilon \right)\!\!\left(2\sigma_{{\bm{\hat{Y}}_t^i}\bm{Y}_t^i} \!+\! \epsilon^{\prime}\right)}{\left(\mu_{\bm{\hat{Y}}_t^i}^2 \!+\! \mu_{\bm{Y}_t^i}^2 \!+\! \epsilon\right)\!\!\left(\sigma_{\bm{\hat{Y}}_t^i}^2 \!+\! \sigma_{\bm{Y}_t^i}^2 \!+\! \epsilon^{\prime}\right)},  
\end{align}
where $\bm{\hat{Y}^i}$ denotes the $i^\text{th}$~predicted \sits with $C$ spectral bands and $T^i$ frames, $\bm{Y}^i$ the corresponding ground truth \sits, and \mbox{$\left (x, y \right )$} and $t$ the spatial and temporal coordinates. In the SSIM computation, $\mu_{\bm{\hat{Y}}_t^i}$ and $\sigma_{\bm{\hat{Y}}_t^i}$ express the mean reflectance and standard deviation of $\bm{\hat{Y}}^i$ at time step $t$, $\mu_{\bm{Y}_t^i}$ and $\sigma_{\bm{Y}_t^i}$ the corresponding mean and standard deviation w.r.t.\ $\bm{Y}^i$, and $\sigma_{{\bm{\hat{Y}}_t^i}\bm{Y}_t^i}$ their covariance. $\epsilon$ and $\epsilon^{\prime}$ are small constants for numerical stability.

MAE, RMSE, and PSNR are popular metrics that quantify the pixel-wise reconstruction error. MAE and RMSE are expressed relative to the TOA reflectance range ($\rho_{\text{TOA}}$, recall that reflectances have been rescaled $\rho_\text{TOA}\in[0, 1]$). PSNR is in decibel (dB). SAM measures the spectral fidelity of the reconstructed pixels, defined as the average angle (in degrees) between predicted and ground truth spectral vectors. SSIM is a unitless per-image metric that measures the overall structural similarity between a reconstructed image and the corresponding ground truth. We compute the pixel-based metrics over all imputed pixels (according to the input mask). Similarly, to compute the average SSIM, we only take images into account that contain masked pixels, denoted as $T_m^i$.

To generate the cloud-free reference \sits for evaluation, we define a global threshold on the cloud probability maps (if available) or the binary cloud masks. That threshold may not be ideal for every single image; consequently, small clouds or haze may remain undetected. To alleviate the impact of these remaining data gaps (in the ground truth) on quantitative analysis, we compute the pixel-based metrics only for those imputed pixels \mbox{$\left (t, x, y \right )$} for which a cloud-free reference reflectance is available according to the original cloud masks, denoted as $\Omega^i$ for the $i^\text{th}$~\sits.

\subsection{Baseline methods}
We compare \utilise against several natural baselines that operate independently on every spatial location over time. The simplest baseline (\emph{last}) imputes missing pixels by copying the last valid observation before the current frame. The next baseline (\emph{closest}) copies either the last or the next following observation, depending on which one is closer in time. The third baseline linearly interpolates between the last and next observations (according to the absolute time span in days), an approach frequently employed in operational practice~\cite{inglada2015assessment}.

\section{Results}
\label{sec:results}
\begin{table*}[!ht]
    \caption{Quantitative comparison of different imputation methods for time series of the \earthnet dataset, separately evaluated for geographic locations similar to the training set (\textit{iid}) and for previously unseen locations (\textit{ood}). Columns~3--7 report metrics computed over all pixels (or images, in the case of SSIM) with missing input data, whereas columns~8--9 measure the quality of output pixels (or images) with valid input values.}
    \centering
	\begin{adjustbox}{max width=0.95\textwidth}
        \begin{tabular}{@{\hspace{1mm}}l|l|ccccc|cc@{\hspace{1mm}}}
            \toprule
            Test split & Method & $\downarrow$ MAE [$\rho_{\text{TOA}}$] & $\downarrow$ RMSE [$\rho_{\text{TOA}}$] & $\downarrow$ SAM [\textdegree] & $\uparrow$ PSNR [dB] & $\uparrow$ 
 SSIM [-] & $\downarrow$ MAE [$\rho_{\text{TOA}}$] & $\uparrow$ 
 SSIM [-] \\           
            \midrule
            \multirow{4}{*}{iid} & Last & 0.0148 & 0.0234 & 3.17 & 33.90 & 0.944 & \textbf{0.0000} & \textbf{1.000} \\
            & Closest                   & 0.0128 & 0.0203 & 2.74 & 35.01 & 0.953 & \textbf{0.0000} & \textbf{1.000} \\
            & Linear interpolation      & 0.0110 & 0.0172 & 2.35 & 36.48 & 0.962 & \textbf{0.0000} & \textbf{1.000} \\
            & \utilise (ours)           & \textbf{0.0086} & \textbf{0.0140} & \textbf{1.87} & \textbf{38.29} & \textbf{0.970} & 0.0002 & \textbf{1.000} \\
            \midrule
            \multirow{4}{*}{ood} & Last & 0.0174 & 0.0270 & 3.36 & 33.13 & 0.936 & \textbf{0.0000} & \textbf{1.000} \\
            & Closest                   & 0.0147 & 0.0230 & 2.84 & 34.31 & 0.948 & \textbf{0.0000} & \textbf{1.000} \\
            & Linear interpolation      & 0.0129 & 0.0197 & 2.42 & 35.67 & 0.957 & \textbf{0.0000} & \textbf{1.000} \\
            & \utilise (ours)           & \textbf{0.0110} & \textbf{0.0170} & \textbf{2.07} & \textbf{36.95} & \textbf{0.964} & 0.0002 & \textbf{1.000} \\
            \bottomrule
        \end{tabular}
    \end{adjustbox}
    \label{tab:results_main}
\end{table*}

\begin{figure*}[!ht]
    \def\mywidth{0.95\textwidth}
    \centering
    \begin{tabular}{@{} m{\mywidth} @{}}
        \includegraphics[trim={0.22cm 0.32cm 0.2cm 0.2cm},clip,width=\mywidth]{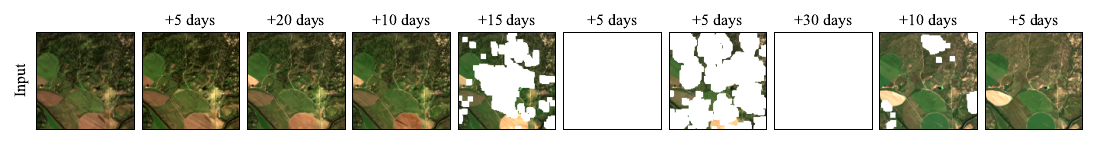}
        \\
        \includegraphics[trim={0.22cm 0.32cm 0.2cm 0.2cm},clip,width=\mywidth]{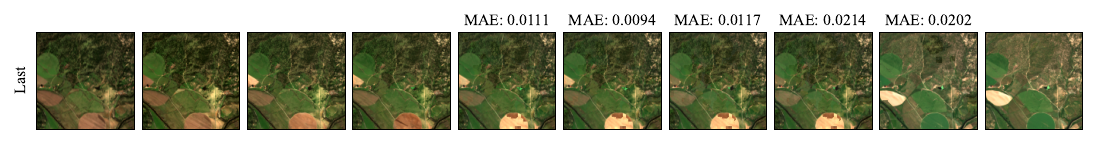}
        \\
        \includegraphics[trim={0.22cm 0.32cm 0.2cm 0.2cm},clip,width=\mywidth]{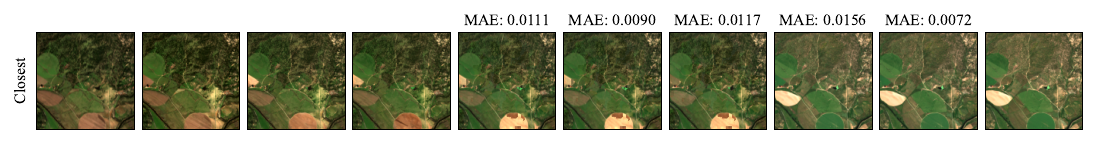}
        \\
        \includegraphics[trim={0.22cm 0.32cm 0.2cm 0.2cm},clip,width=\mywidth]{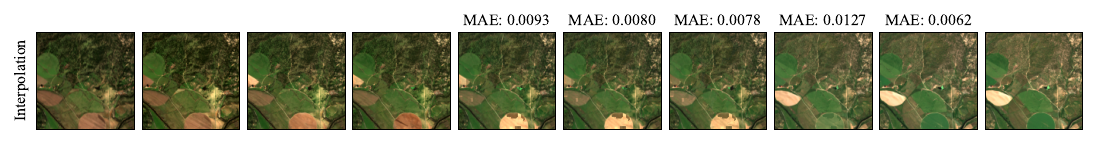}
        \\
        \includegraphics[trim={0.22cm 0.32cm 0.2cm 0.2cm},clip,width=\mywidth]{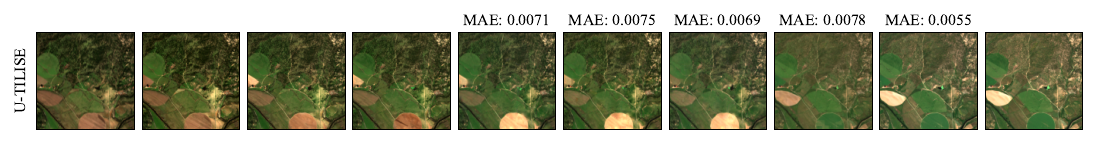}
        \\
        \includegraphics[trim={0.22cm 0.32cm 0.2cm 0.48cm},clip,width=\mywidth]{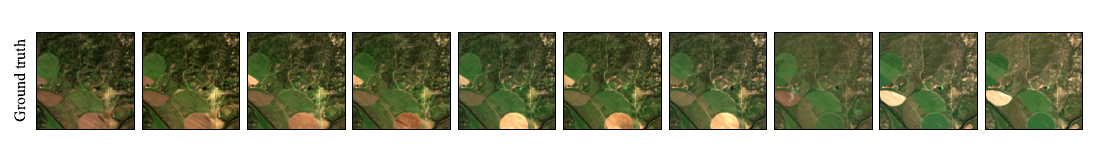}
        \\
    \end{tabular}
    \caption{Visual comparison of \utilise with selected baselines for a \sentinelII time series of the \earthnet \textit{iid}~test split with predominantly gradually changing land cover. We show the true-color RGB composite for every image in the time series. The number above an image denotes its temporal distance to the previous observation in the sequence or its MAE, evaluated over all pixels that have been masked in the corresponding input image and across all spectral bands (R, G, B, and NIR).}
    \label{fig:fields}
\end{figure*}
\begin{figure*}[!ht]
    \def\mywidth{0.95\textwidth}
    \centering
    \begin{tabular}{@{} m{\mywidth} @{}}
        \includegraphics[trim={0.22cm 0.32cm 0.2cm 0.2cm},clip,width=\mywidth]{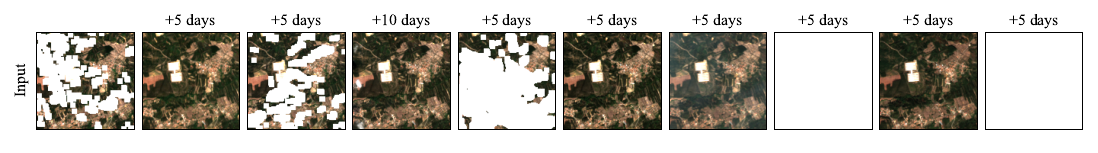}
        \\
        \includegraphics[trim={0.22cm 0.32cm 0.2cm 0.2cm},clip,width=\mywidth]{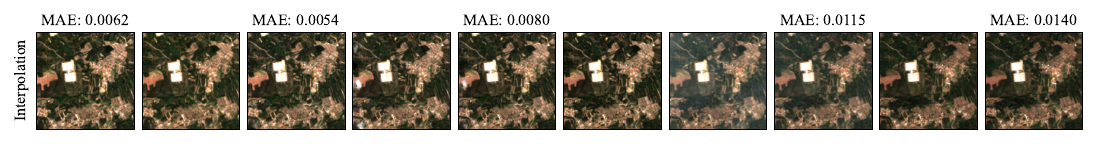}
        \\
        \includegraphics[trim={0.22cm 0.32cm 0.2cm 0.2cm},clip,width=\mywidth]{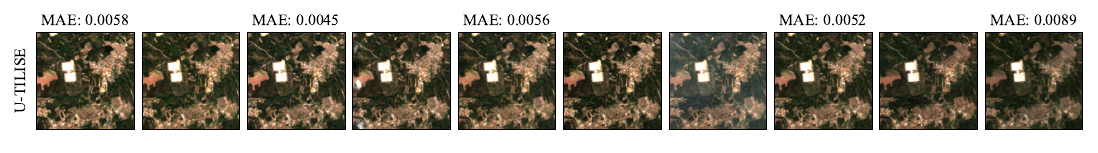}
        \\
        \includegraphics[trim={0.22cm 0.32cm 0.2cm 0.48cm},clip,width=\mywidth]{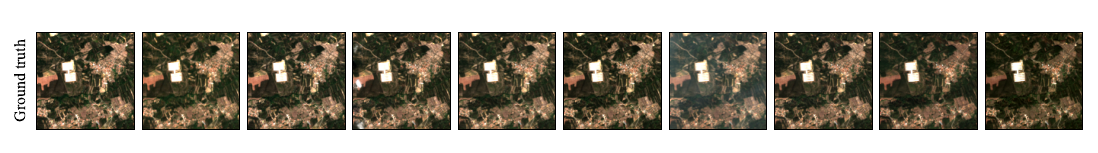}
        \\
        \smallskip
        \\
        \includegraphics[trim={0.22cm 0.32cm 0.2cm 0.2cm},clip,width=\mywidth]{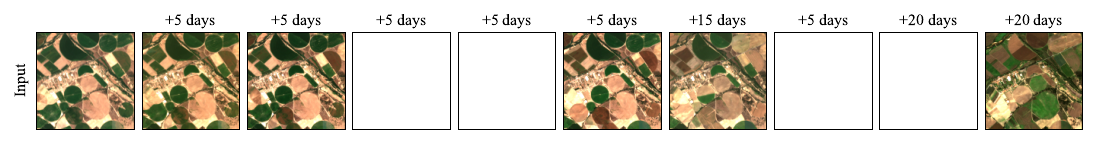}
        \\
        \includegraphics[trim={0.22cm 0.32cm 0.2cm 0.2cm},clip,width=\mywidth]{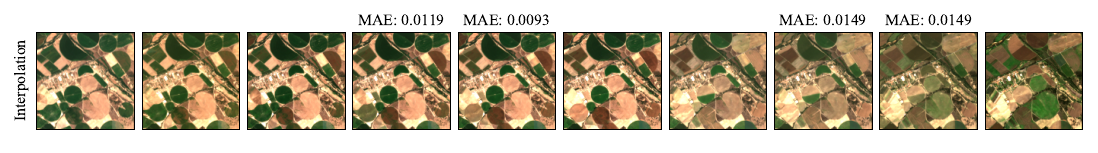}
        \\
        \includegraphics[trim={0.22cm 0.32cm 0.2cm 0.2cm},clip,width=\mywidth]{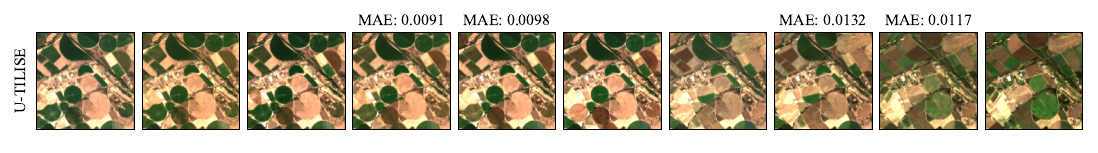}
        \\
        \includegraphics[trim={0.22cm 0.32cm 0.2cm 0.48cm},clip,width=\mywidth]{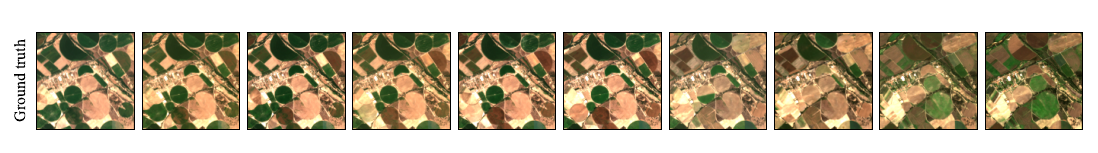}
        \\
    \end{tabular}
    \caption{Visual comparison (true-color RGB composites) of two \sentinelII time series of the \earthnet \textit{iid}~test split, gap-filled using either \utilise or linear interpolation over time. Rows~1--4 depict a static scene, while rows~5--8 show a scene with sudden land cover changes due to agricultural activities by humans. The number above an image indicates its temporal distance to the previous observation in the sequence or its MAE, evaluated over all pixels that have been masked in the corresponding input image and across all spectral bands (R, G, B, and NIR).}
    \label{fig:static_sudden}
\end{figure*}

\subsection{Imputation quality of \utilise}
We begin by evaluating our method on \sentinelII time series augmented with synthetic data gaps. Quantitative results are shown in Table~\ref{tab:results_main}, visual examples are given in Figs.~\ref{fig:fields} and~\ref{fig:static_sudden}. \utilise generates coherent and gap-free optical time series that capture the natural evolution of the depicted land cover. It can handle occlusions of various shapes and sizes, recover images that suffer from severe occlusions, and reconstruct complete times series from input sequences that include multiple consecutive frames of missing data. Note how the model has implicitly learned to adapt to radiometric variations within the time series,%
\footnote{Primarily caused by illumination changes due to atmospheric effects.}
such that imputed pixels seamlessly blend into the surrounding image content, even if the frames that show the same regions unoccluded are radiometrically different. Besides naturally adapting reflectance values, \utilise is able to recover plausible color transitions and impute missing information in scenes with non-trivial temporal dynamics (Fig.~\ref{fig:fields}, row~5). Furthermore, it preserves high reflectance values not associated with clouds (Fig.~\ref{fig:static_sudden}, row~3). Abrupt scene changes, such as harvested agricultural fields, are sometimes missed (Fig.~\ref{fig:static_sudden}, row~7), which is natural since their exact timing depends on weather conditions that the model has no access to.

\begin{figure*}[!th]
    \centering
    \def\myheight{0.219\textwidth}
    \includegraphics[trim={0.2cm 0.25cm 2.2cm 0.85cm},clip,height=\myheight]{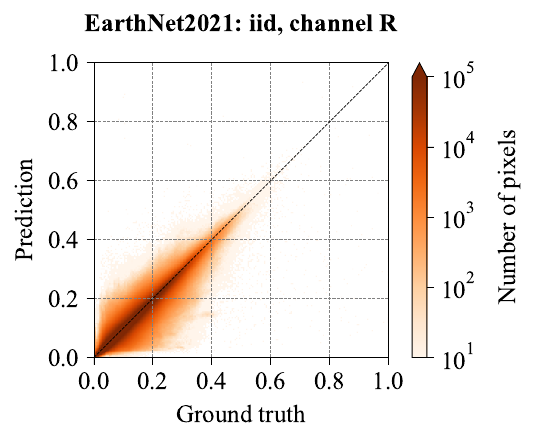}
    \includegraphics[trim={0.2cm 0.25cm 2,2cm 0.85cm},clip,height=\myheight]{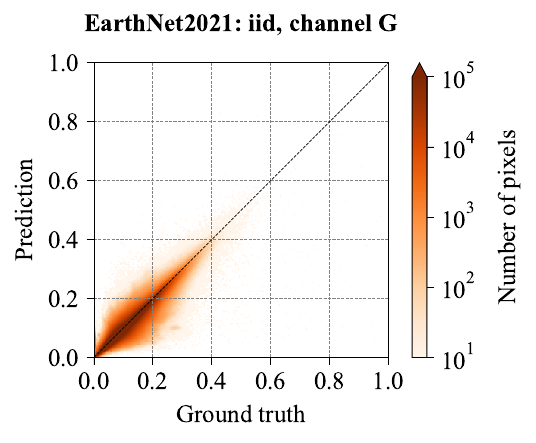}
    \includegraphics[trim={0.2cm 0.25cm 2.2cm 0.85cm},clip,height=\myheight]{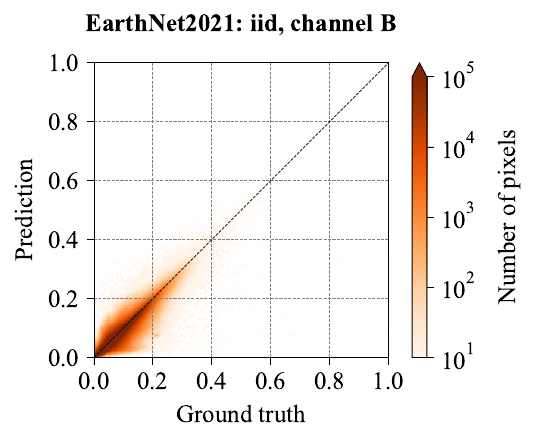}
    \includegraphics[trim={0.2cm 0.25cm 0.2cm 0.85cm},clip,height=\myheight]{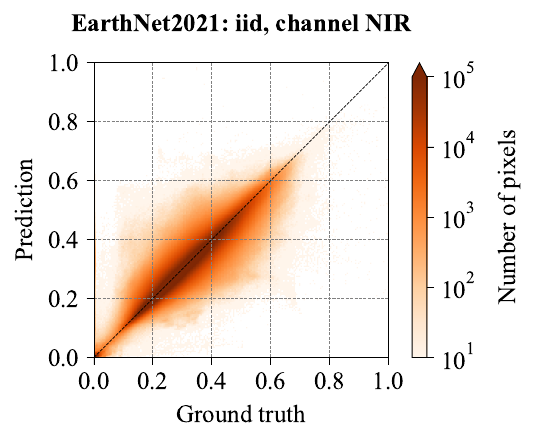}
    \caption{Channel-wise ground truth reflectances versus reflectances regressed by \utilise for missing pixels in optical time series of the \earthnet dataset (\textit{iid}~split). From left to right: B4~(red), B3~(green), B2~(blue), and B8~(near-infrared).}
    \label{fig:scatter_plot}
\end{figure*}

Quantitatively, the predicted reflectances agree well with the true values across all optical bands. See Fig.~\ref{fig:scatter_plot}. \utilise yields a MAE below 1\% of the intensity range and a PSNR on the order of 38$\,$dB. Recall, these errors are averaged only over imputed pixels and not inflated by the trivial reconstruction of observed values (cf.\ Table~\ref{tab:results_main}, \textit{iid}~test split). The SSIM, averaged over all images with data gaps, amounts to 0.97. Notably, there is only a moderate performance penalty when applying the model to previously unseen locations (cf.\ Table~\ref{tab:results_main}, \textit{ood}~test split). The MAE increases by 27\% to roughly 0.01 and the SAM by 10\% from 1.9$\,$degrees to 2.1$\,$degrees. Yet, with a PSNR of almost 37$\,$dB and a SSIM of 0.96, this still amounts to high fidelity and visual quality.

\subsection{Comparison to baselines}
Table~\ref{tab:results_main} compares \utilise against the three baselines. We first discuss the performance on \textit{iid}~test sequences, corresponding to the imputation of new sequences acquired at previously seen locations. As expected, the \emph{last} baseline, which copies the last valid observation, yields the largest reconstruction errors. Instead, cloning the temporally \emph{closest} observation improves the reconstruction quality and the spectral fidelity markedly, reducing MAE, RMSE, and SAM by more than 10\% and increasing PSNR by about 1$\,$dB; mostly because the \emph{last} heuristic degrades for longer data gaps where the same location is repeatedly occluded.
\emph{Linear interpolation} between the most recent and the next available observation brings further gains. MAE, RMSE, and SAM drop by another $\approx\,$14\%, and PSNR improves by 1.5$\,$dB to 36.0$\,$dB. \utilise consistently outperforms all baselines by a significant margin. Compared to linear interpolation, the MAE, RMSE, and SAM values decrease by 20\%, and the PSNR increases by more than 1.5$\,$dB. We observe similar trends when evaluating the \textit{ood}~test set of previously unseen locations. We note in 
passing that the \textit{ood} test set is objectively more difficult: all methods perform slightly worse on it, although the baselines do not involve any learning and can, by definition, not overfit to specific geographic locations.

\utilise predicts a reflectance value for \emph{every} spatio-temporal location in the output sequence, including pixels with valid input observations. In principle, those predicted values could deviate from the actual, observed values at cloud-free pixels.\footnote{Obviously, one could, in practice, copy valid input pixels to the output. We regard this as a postprocessing option available to any cloud removal method, not as a part of the actual model.} For a complete evaluation, we thus also evaluate the spectral fidelity at pixels with valid input reflectances (Table~\ref{tab:results_main}, columns~8--9). By construction, all baselines achieve the same, maximal performance, since they do not alter valid reflectance values. \utilise, on the other hand, must learn to preserve the reflectances at unoccluded (spatio-temporal) locations. It does that astonishingly well, with a MAE around $\frac{1}{5000}$ of the intensity range---less than the radiometric sensitivity of \sentinelII.\footnote{\url{https://sentinels.copernicus.eu/web/sentinel/user-guides/sentinel-2-msi/resolutions/radiometric}}

The estimates of \utilise are also qualitatively superior to those of the baselines, especially in the presence of significant spectral changes in time (cf.\ Fig.~\ref{fig:fields}) and of remaining atmospheric effects (cf.\ Fig.~\ref{fig:static_sudden}, rows~1--4). A striking example is the southernmost, circular field in Fig.~\ref{fig:fields}, for which \utilise smoothly transitions from dark to bright brown and then changes abruptly to dark green, which agrees well with the true evolution of the depicted scene. In contrast, copying or simple interpolation lead to evident visual artifacts.

\begin{figure}[!t]
    \centering
    \includegraphics[trim={0.33cm 0.3cm 0.2cm 0.2cm},clip,width=\columnwidth]{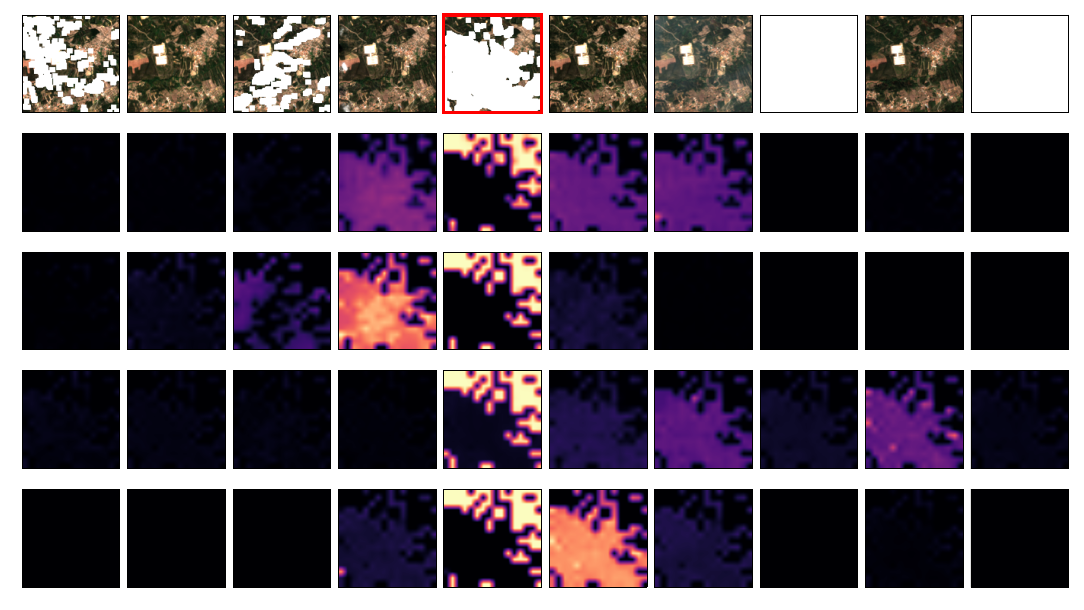}
    \caption{Self-attention in the temporal encoder. We show the attention scores for imputing the \nth{5} image in the example time series (highlighted with a red frame in row~1), displayed separately for each of the four attention heads (rows~2--5). The attention masks are bilinearly upsampled to the native spatial resolution of the input time series and color-coded from black (no attention) to yellow (maximum attention).}
    \label{fig:attention_across_heads}
\end{figure}

\begin{figure}[!th]
    \centering
    \includegraphics[trim={0.33cm 4.3cm 3.75cm 0.2cm},clip,width=\columnwidth]{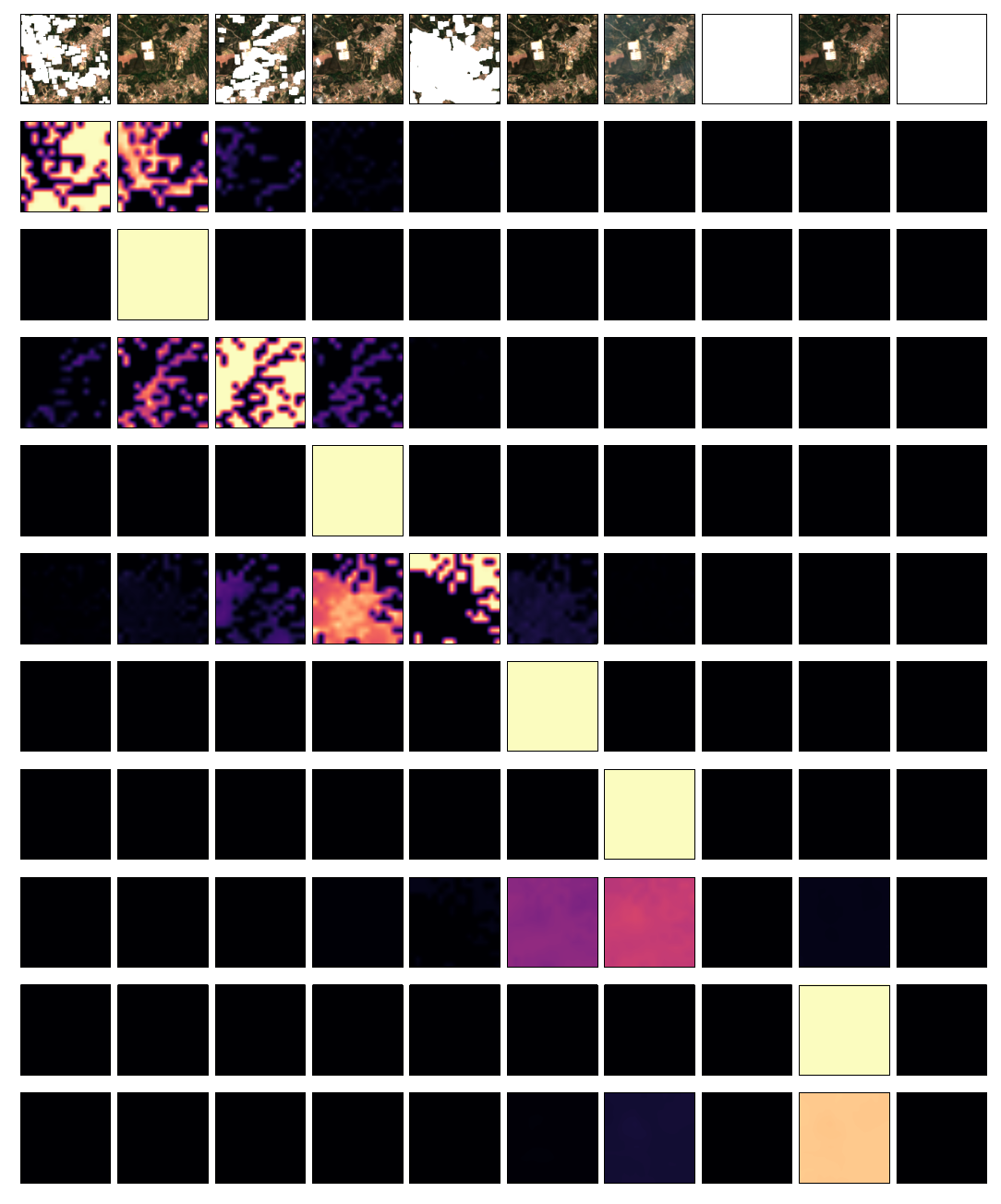}
    \caption{Self-attention in the temporal encoder. For the time series in row~1. Rows~2--9 show the attention masks for one of the four heads, with rows corresponding to frames of the output (in temporal order). The masks are bilinearly upsampled to the native spatial resolution of the input and color-coded from black (no attention) to yellow (maximum attention). Note how the attention progressively moves through time to focus on unoccluded inputs and unoccluded pixels of partially occluded inputs, while it borrows information from temporally nearby frames where needed.}
    \label{fig:attention_across_time}
\end{figure}

\subsection{Learned multi-temporal attention}
The attention masks of the temporal self-attention mechanism encode the contribution of each input pixel to the regression of the output pixels.%
\footnote{Strictly speaking, the attention scores express the contribution w.r.t.\ the high-dimensional, but spatially coarsened latent embedding.}
This makes them a useful visual cue to determine on which input frames the learned model bases its predictions. Figs.~\ref{fig:attention_across_heads} and~\ref{fig:attention_across_time} depict the attention masks for the first time series in Fig.~\ref{fig:static_sudden}, demonstrating that \utilise indeed discovers, in a data-driven manner, which observations are most useful for its task. Not only is the attention low in data gaps (cf.\ Fig.~\ref{fig:attention_across_time}), the model has also learned to preferentially attend to temporally close observations; to use information from unoccluded regions of the current frame, presumably to match its radiometry; and to let the attention heads specialize on different portions of the sequence (cf.\ Fig.~\ref{fig:attention_across_heads}).

\begin{figure}[!t]
    \centering
    \subfloat[]{\includegraphics[trim={0.35cm 0.33cm 0.22cm 0.2cm},clip,scale=1.2]{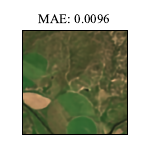}}
    \hfill
    \subfloat[]{\includegraphics[trim={0.35cm 0.33cm 0.22cm 0.2cm},clip,scale=1.2]{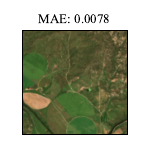}}
    \hfill
    \subfloat[]{\includegraphics[trim={0.35cm 0.33cm 0.22cm 0.2cm},clip,scale=1.2]{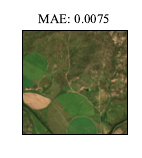}}
    \hfill
    \subfloat[]{\includegraphics[trim={0.35cm 0.33cm 0.22cm 0.2cm},clip,scale=1.2]{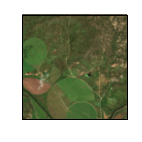}}
    \smallskip
    \caption{Visual comparison of different \utilise variants for the \nth{8} time step in the sequence from Fig.~\ref{fig:fields} (the \nth{2} totally masked image). (a)~\utilise with ordinary skip connections; (b)~\utilise with default parameter settings; (c)~\utilise with 16~attention heads; (d)~ground truth.}
    \label{fig:ablations}
\end{figure}

\subsection{Importance of temporal encoding}
\begin{table*}[!t]
    \caption{Quantitative results of different \utilise variants. We ablate the influence of the temporal encoder and temporally weighted skip connections in the spatial U-Net (rows~2--3), the strategy used for positional encoding (rows~4--6), the number of images~$T$ processed in one shot (rows~7--8), and the number of attention heads~G of the temporal encoder (rows~9--11). The metrics are computed over all pixels (or images, in the case of SSIM) with missing data in the input time series.}
    \centering
	\begin{adjustbox}{max width=0.95\textwidth}
        \begin{tabular}{ccccc|ccccc}
            \toprule
            Temp. attention & Weighted skip & Positional encoding & $T$ & $G$ & $\downarrow$ MAE [$\rho_{\text{TOA}}$] & $\downarrow$ RMSE [$\rho_{\text{TOA}}$] & $\downarrow$ SAM [\textdegree] & $\uparrow$ PSNR [dB] & $\uparrow$ SSIM [-] \\       
            \midrule
            \checkmark & \checkmark & day within year & 10 & 4 & 0.0100 & 0.0160 & 2.16 & 36.82 & 0.964 \\
            \checkmark &  & day within year           & 10 & 4 & 0.0120 & 0.0187 & 2.75 & 35.14 & 0.936 \\
            & & day within year                       & 10 & 4 & 0.0339 & 0.0498 & 8.57 & 26.49 & 0.806 \\
            \midrule
            \checkmark & \checkmark & day within sequence & 10 & 4 & 0.0098 & 0.0158 & 2.07 & 37.00 & \textbf{0.965} \\
            \checkmark & \checkmark & enumeration         & 10 & 4 & 0.0124 & 0.0199 & 2.61 & 34.93 & 0.953 \\	
            \checkmark & \checkmark & -                   & 10 & 4 & 0.0142 & 0.0228 & 3.14 & 33.91 & 0.947 \\
            \midrule
            \checkmark & \checkmark & day within year & 5 & 4  & 0.0102 & 0.0163 & 2.18 & 36.66 & 0.963 \\
            \checkmark & \checkmark & day within year & 15 & 4 & 0.0099 & 0.0159 & 2.14 & 36.89 & 0.964 \\
            \midrule
            \checkmark & \checkmark & day within year & 10 & 2  & 0.0101 & 0.0162 & 2.18 & 36.77 & 0.963 \\
            \checkmark & \checkmark & day within year & 10 & 8  & 0.0098 & 0.0158 & 2.12 & 36.99 & 0.964 \\
            \checkmark & \checkmark & day within year & 10 & 16 & \textbf{0.0097} & \textbf{0.0156} & \textbf{2.10} & \textbf{37.09} & \textbf{0.965} \\
            \bottomrule
        \end{tabular}
        \end{adjustbox}
    \label{tab:results_ablations}
\end{table*}

We go on to study the influence of different network configurations, starting with the temporal encoder. To this end, we conduct two ablation experiments: \textit{(i)} we replace the temporally weighted skip connections between corresponding layers of the spatial encoder and decoder with ordinary skip connections, and \textit{(ii)} we remove the temporal encoder altogether, resulting in a standard U-Net architecture that processes each frame of the input sequence independently. As expected, we observe a significant deterioration in all evaluation metrics when \utilise cannot exploit the time series to fill in missing image content (Table~\ref{tab:results_ablations}, \nth{3}~row): without a mechanism to represent spatio-temporal relations, the model must hallucinate the missing spectral values only from the unoccluded context in the same image. In the extreme case of a fully occluded view, this means blindly synthesizing a multi-spectral image from a blank slate. Due to this severe ill-conditioning, the model merely learns to reproduce observed values while replacing missing pixels with the average reflectance value of the training set.
On the contrary, as soon as the temporal encoder is added in the bottleneck of the U-Net, while still using ordinary skip connections otherwise, \utilise learns to leverage the temporal context to inform the reconstruction of missing pixels. MAE, RMSE, and SAM are about 3$\times$ lower than without temporal encoding, while the PSNR increases by 8.7$\,$dB (Table~\ref{tab:results_ablations}, \nth{2}~row). Visually, \utilise predicts plausible spectral intensities and colors in occluded image regions but fails to recover fine-grained texture details (Fig.~\ref{fig:ablations}(a)). The full model, which also uses the upsampled attention maps in the skip connections, manages to restore visibly more high-frequency details (Fig.~\ref{fig:ablations}(b)). The relative improvement amounts to $\approx\,$15\% in MAE and RMSE, and more than 20\% in terms of SAM (Table~\ref{tab:results_ablations}, \nth{1}~row). PSNR improves by another 1.7$\,$dB.

\begin{figure}[!t]
    \centering
    \includegraphics[trim={0.25cm 0.25cm 0.22cm 0.2cm},clip,width=\columnwidth]{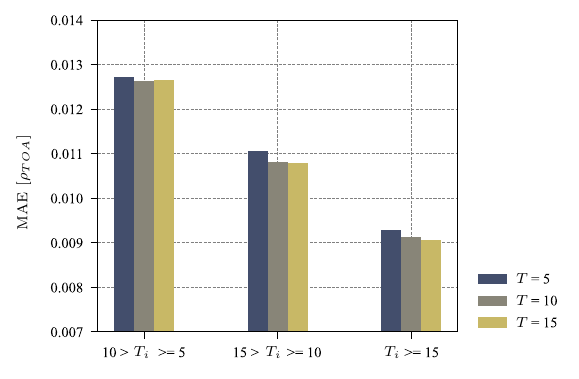}
    \caption{MAE as a function of the full sequence length $T_i$ and the temporal window~$T$ of \utilise.}
    \label{fig:Ti_vs_T}
\end{figure}

\subsection{Choice of positional encoding}
Next, we evaluate the influence of the positional encoding scheme (cf.\ Section~\ref{sec:pe}) on the model output. We found experimentally that information about the image order and the relative temporal distance between observations is crucial to reconstruct realistic, gap-free time series. Without any positional encoding, the performance of \utilise  drops significantly; MAE, RMSE, and SAM increase by $\approx\,$40\% and PSNR drops by almost 3$\,$dB (Table~\ref{tab:results_ablations}, \nth{6}~row). Injecting ordinal information improves MAE and RMSE by 13\% and SAM by 17\%, while also improving the perceptual similarity to the target sequence (Table~\ref{tab:results_ablations}, \nth{5}~row). Encoding the temporal offset from the first observation in the sequence rather than the ordinal position brings another improvement of $\approx\,$20\% in MAE, RMSE, and SAM and boosts PSNR by 2$\,$dB (Table~\ref{tab:results_ablations}, \nth{4}~row). We find only a tiny difference between encoding the temporal distance to the first observation or encoding the acquisition date of every observation relative to the \nth{1} January of the respective calendar year. We speculate that the latter strategy suffers from a bias inherent in our training data: most of the \earthnet time series are captured between May and October; likely, the dataset does not offer sufficient variability to extract an expressive prior over seasonal patterns.

\begin{figure*}[!ht]
    \def\mywidth{0.971\textwidth}
    \setlength{\tabcolsep}{0.2em}
    \centering
    \begin{tabular}{l m{\mywidth}}
        \rotatebox[origin=c]{90}{\footnotesize \hspace{0.1cm}Observed} & \includegraphics[trim={0.35cm 0.35cm 0.23cm 0.22cm},clip,width=\mywidth]{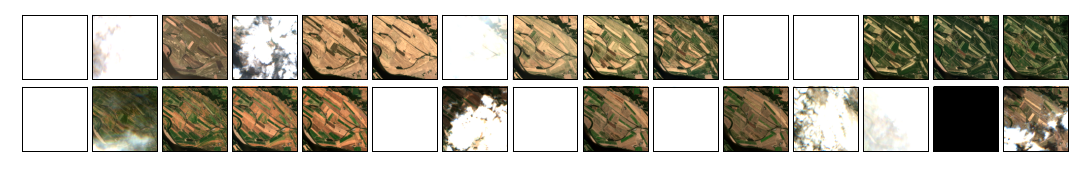}
        \\
        \vspace{-0.5em}
        \\
        \rotatebox[origin=c]{90}{\footnotesize \hspace{0.2cm}\utilise} & \includegraphics[trim={0.35cm 0.35cm 0.23cm 0.22cm},clip,width=\mywidth]{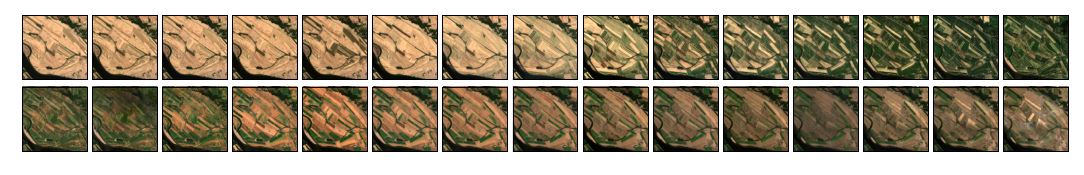}
        \\
        \smallskip
        \\
        \rotatebox[origin=c]{90}{\footnotesize \hspace{0.1cm}Observed} & \includegraphics[trim={0.35cm 0.35cm 0.23cm 0.22cm},clip,width=\mywidth]{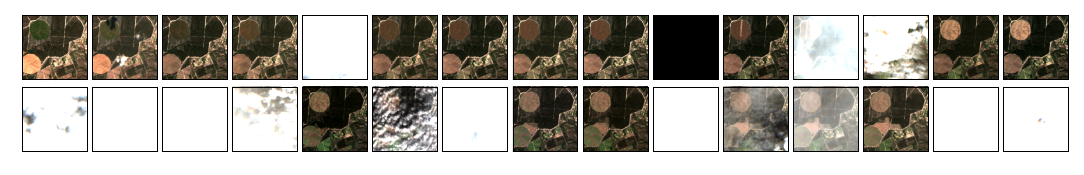}
        \\
        \vspace{-0.5em}
        \\
        \rotatebox[origin=c]{90}{\footnotesize \hspace{0.2cm}\utilise} & \includegraphics[trim={0.35cm 0.35cm 0.23cm 0.22cm},clip,width=\mywidth]{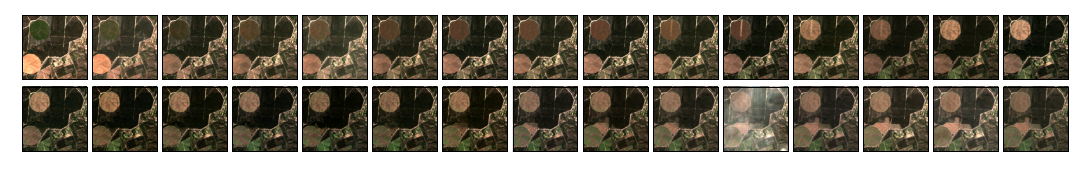}
        \\
    \end{tabular}
    \vspace{-0.2em}
    \caption{Imputation results for two time series of the \earthnet dataset with real data gaps due to clouds, cloud shadows, and missing frames (shown in black). Example~1 is from the \textit{ood}~test split and example~2 from the \textit{iid}~test split.}
    \label{fig:real_clouds}
\end{figure*}

\subsection{Influence of input sequence length $T$}
To test the sensitivity of \utilise to the number of images processed as one sequence, we define two model variants by varying the temporal window length $T$. In detail, we retrain \utilise once with training sequences that are randomly trimmed to a maximum temporal length of five images (\mbox{$T=5$}), and once with time series that comprise up to 15~images  (\mbox{$T=15$}). At test time, we always process the full-length time series, employing a sliding window scheme if the temporal length $T_i$ is larger than the maximum length $T$ used during training.

We find that all evaluation metrics remain relatively stable when varying the temporal window of \utilise (Table~\ref{tab:results_ablations}, rows~7--8), with fluctuations below 2\% compared to our default setting (\mbox{$T=10$}). For a more fine-grained analysis, we thus separately measure the performance for time series consisting of (\textit{i})~at most 9~images, (\textit{ii})~10 to 14~images, or (\textit{iii})~15 or more images. As shown in Fig.~\ref{fig:Ti_vs_T}, the MAE at imputed pixels slightly reduces with longer time series and temporal window $T$, indicating that \utilise can exploit long temporal context if needed.

\subsection{Number of attention heads}
\utilise is fairly robust with varying numbers of temporal attention heads. We find marginal quantitative gains when adding more heads (Table~\ref{tab:results_ablations}, rows~9-11), and also only small differences in visual quality (Fig.~\ref{fig:ablations}(c)).

\subsection{Time series with real data gaps}
In the last experiment, we use the trained \utilise model, unaltered, to impute missing pixels in time series with real data gaps. This scenario corresponds to the practical application case, where the masked pixels are truly unobserved. Of course, this also implies that the resulting outputs can only be assessed through visual inspection, since no ground truth exists for the masked areas. 
Note that the application to real cloudy time series may, from a machine learning perspective, involve some degree of generalization. As clouds do not have sharp boundaries, the unmasked regions just outside the cloud mask may, in some cases, still be affected by thin clouds and haze. During training, where cloud-free images are synthetically masked, the model has not been exposed to such a situation.

Fig.~\ref{fig:real_clouds} depicts imputation results for two representative \mbox{30-frame} time series from the \earthnet test set. The original, observed time series suffer from severe data gaps from clouds, shadows, haze, and missing images. Furthermore, the second example exhibits a rather long period without any valid observations. Despite these challenges, \utilise creates realistic, gap-free time series of high visual quality. The reconstructed time series do occasionally suffer from remaining clouds or haze, if they were missed by the cloud masking algorithm (Fig.~\ref{fig:real_clouds}, \nth{5} last frame of the second example).

\section{Conclusion}
\label{sec:conclusion}
We have presented \utilise, a learned sequence-to-sequence model for data imputation in optical satellite image time series. The model combines 2D~convolutions over the spatial and spectral dimensions and 1D~self-attention across time into an efficient prior over multi-spectral and multi-temporal reflectance patterns. Given an optical time series in which invalid reflectance values are masked, \utilise creates a coherent time series with a clean, cloud-free image at every time step of the input.

In a series of experiments, we have shown that \utilise reconstructs gap-free \sentinelII sequences with high accuracy. It removes clouds and cloud shadows of various shapes and sizes, manages to recover multiple consecutive frames of missing data, and generalizes to previously unseen geographical locations. On the \earthnet dataset, the average MAE within the data gaps is in the order of 1\% of the intensity range, and the PSNR is $\approx\,$38$\,$dB.

A limitation of the current approach is that it relies on cloud (and cloud shadow) masks as auxiliary input. Detecting cloudy pixels in remotely sensed imagery is challenging and still not completely solved. Mistakes of the preceding cloud detector limit the performance of \utilise, since it implicitly learns to preserve input reflectances that are valid according to the masks. Furthermore, the model may of course reconstruct radiometrically plausible but incorrect transitions in cases where the available information is too sparse to determine when a sudden change has occurred, such as, for instance, a harvesting event during a multi-frame data gap.

An interesting future research direction is to eliminate the need for external cloud masks and instead design the model such that it implicitly performs the detection of invalid pixels. Another useful extension for users of the \utilise output will be to integrate a probabilistic deep learning scheme and supplement the output with spatially and temporally resolved uncertainty estimates.

\appendix

\section*{Complementary experiments}
\label{sec:appendix}
Some authors~\cite{sebastianelli2022plfm,ebel2022sen12ms} have advocated the use of co-registered SAR observations, which are largely unaffected by the atmosphere, to support optical image imputation. It is a natural idea to also augment \utilise with a time series of spatially and temporally co-registered SAR images. Technically, this is straightforward: we add the \mbox{two-channel} SAR images (ortho-rectified VV/VH log-amplitude) as additional input channels that need not be reconstructed and, accordingly, increase the filter depth of the first layer by two. We train and test this multi-modal variant of \utilise in a simulated setting using the \senXIImscrts dataset~\cite{ebel2022sen12ms}, a multi-modal and multi-temporal dataset specifically designed for multi-modal cloud removal. Unfortunately, it turns out that the dataset is not only considerably smaller than \earthnet (see Table~\ref{tab:datasets}), but its sequences also exhibit comparatively lower temporal variability and dynamics and do not allow a conclusive comparison. We refrain from drawing firm conclusions about the impact of complementary SAR observations and instead conclude that a more informative dataset must be created to investigate the issue, possibly by augmenting \earthnet with SAR observations.

\subsection*{\senXIImscrts dataset}
\senXIImscrts~\cite{ebel2022sen12ms} comprises about $15\,000$~globally sampled \sentinelII \sits (Level-1C with top-of-atmosphere reflectances) from 2018 with a spatial extent of 256$\times$256$\,$pixels (2.56$\times$2.56$\,$km in scene space). Each time series contains 30~images, with varying temporal spacing of 5 to 15~days between consecutive observations. The images encompass all 13~spectral bands, upsampled to 10$\,$m resolution. Furthermore, every optical image is paired with a spatially co-registered, temporally close (but not synchronous) \mbox{C-band} SAR image with two channels representing the $\sigma_0$ back-scatter coefficients in the VV and VH polarizations, in units of decibels (dB). The temporal offset between paired optical and SAR observations is three days on average, although, for 5.5\% of the pairs, the temporal difference is over one week. The dataset also includes pixel-wise cloud probabilities and binary cloud masks, produced with the \texttt{S2Cloudless}  detector~\cite{baetens2019validation}.
 
\begin{table}[!t]
    \centering
    \caption{Acquisition details of the two \sentinelII datasets. We list the number of sequences, their average length, and the temporal resolution; separately for the training, validation, and test parts (for \earthnet, further divided into \textit{iid} and \textit{ood}~test sets). We also show the statistics for the simulated training sequences, after removing images with actual clouds.
    }
    \label{tab:datasets}
    \begin{adjustbox}{max width=\columnwidth}
    \begin{tabular}{@{\hspace{\tabcolsep}}lcc@{\hspace{\tabcolsep}}}
        \toprule
        & \earthnet & \senXIImscrts \\
        & \cite{requena2021earthnet2021} & \cite{ebel2022sen12ms} \\
        \midrule
        \multicolumn{3}{@{}l}{\textbf{Acquisition}}\\
        Spatial coverage          & Europe                 & global   \\
        Time period               & Nov 2016--May 2020    & 2018     \\
        Spectral bands            & RGB, NIR               & all 13 bands \\
        Image size [pix]          & 128$\times$128         & 256$\times$256 \\
        Spatial resolution [m]    & 20                     & 10  \\
        Temporal length [-]        & 30                    & 30 \\
        Temporal resolution [days] & regular, 5            & irregular, avg.\ 12 \\
        No. of sequences [-] \\
        \hspace{\tabcolsep} train & $18\,955$                            & $10\,167$ \\
        \hspace{\tabcolsep} val   & \phantom{0}$4\,949$                  & \phantom{0}$1\,410$ \\
        \hspace{\tabcolsep} test  & \phantom{0}$4\,219$ $\mid$ $4\,214$  & \phantom{0}$3\,716$ \\
        \midrule
        \multicolumn{3}{@{}l}{\textbf{Cloud-free subsequences}}\\
        No. of sequences [-] \\
        \hspace{\tabcolsep} train & $18\,823$                & $5\,618$             \\ 
        \hspace{\tabcolsep} val   & \phantom{0}$4\,944$      & \phantom{$0\,$}$937$ \\
        \hspace{\tabcolsep} test  & $3\,948$ $\mid$ $4\,116$ & $2\,238$             \\ 
        Temporal length [-] \\
        \hspace{\tabcolsep} train & 15.1 $\pm$ 4.5  & 11.1 $\pm$ 6.0           \\  
        \hspace{\tabcolsep} val   & 16.4 $\pm$ 5.2  & \phantom{0}7.2 $\pm$ 2.1  \\
        \hspace{\tabcolsep} test  & 13.3 $\pm$ 5.0 $\mid$ 13.8 $\pm$ 5.5 & \phantom{0}8.8 $\pm$ 3.7     \\    
        Temporal resolution [days] \\
        \hspace{\tabcolsep} train & 10.2 $\pm$ 3.3            & 14.9 $\pm$ 2.3 \\ 
        \hspace{\tabcolsep} val   & \phantom{0}9.6 $\pm$ 3.4  & 14.4 $\pm$ 1.9 \\
        \hspace{\tabcolsep} test  & 11.3 $\pm$ 4.6 $\mid$ 11.2 $\pm$ 4.6 & 16.1 $\pm$ 3.1 \\ 
        \bottomrule
    \end{tabular}
    \end{adjustbox}
\end{table}

\subsection*{Experimental setup}
We adopt the preprocessing protocol of~\cite{ebel2022sen12ms,xu2022glf} and value-clip the optical images to the range ${[0, 10\,000]}$ and the SAR images to ${[-25, 0]}$, followed by normalization to the unit range ${[0, 1]}$. Like for \earthnet, we extract cloud-free optical time series for training and evaluation (cf.\ Section~\ref{sec:cloud_filtering}). In rare cases, \senXIImscrts sequences exhibit data gaps of several consecutive months. To limit potential land cover and seasonal changes to a reasonable range, we temporally trim the \sits such that the temporal spacing between adjacent valid frames is at most four weeks. The resulting \sits are, on average, shorter than those of \earthnet, and they have about 50\% larger temporal spacing between consecutive frames (cf.\ Table~\ref{tab:datasets}).

After removing real data gaps, we introduce synthetic gaps into the optical images (cf.\ Section~\ref{sec:simulation}) and concatenate the resulting, masked optical time series with the (unmodified) SAR time series along the channel dimension to produce multi-modal input for \utilise.

\subsection*{Training details}
We use the Adam optimizer~\cite{adam} with hyper-parameters \{\mbox{$\beta_1$=0.9}, \mbox{$\beta_2$=0.999}\}, batch size 3, and a weight decay of~$10^{-5}$. The base learning rate of \mbox{$2 \cdot 10^{-4}$} is reduced by 50\% every 80~training epochs. Due to the larger spatial dimensions of the input time series (256$\times$256$\,$pixels, compared to 128$\times$128$\,$pixels in \earthnet), we add an additional convolutional block in the spatial encoder and decoder, such that the (spatial) dimension of 16$\times$16$\,$pixels in the bottleneck is the same as for \earthnet.

\subsection*{Results}
\begin{table*}[!ht]
    \caption{Quantitative comparison of \utilise with conventional baselines for the \senXIImscrts dataset. The metrics are computed over all pixels (or images, in the case of SSIM) with missing data in the input time series. We train and evaluate \utilise once only with optical time series and once with additional SAR input.}
    \centering
	\begin{adjustbox}{max width=0.95\textwidth}
        \begin{tabular}{@{\hspace{1mm}}l|ccccc@{\hspace{1mm}}}
            \toprule
            Method & $\downarrow$ MAE [$\rho_{\text{TOA}}$] & $\downarrow$ RMSE [$\rho_{\text{TOA}}$] & $\downarrow$ SAM [\textdegree] & $\uparrow$ PSNR [dB] & $\uparrow$ 
 SSIM [-] \\           
            \midrule
            Last                      & 0.0178 & 0.0268 & 5.06 & 32.18  & 0.945 \\
            Closest                   & 0.0165 & 0.0250 & 4.77 & 32.76 & 0.949 \\
            Linear interpolation      & 0.0144 & 0.0218 & 4.22 & 34.02 & 0.957 \\
            \utilise, w/o SAR (ours)  & 0.0149 & 0.0222 & 4.39 & 33.77 & 0.955 \\
            \utilise, w/ SAR (ours)   & 0.0146 & 0.0217 & 4.32 & 33.94 & 0.956 \\
            \bottomrule
        \end{tabular}
    \end{adjustbox}
    \label{tab:results_sar}
\end{table*}

\begin{figure*}[!ht]
    \def\mywidth{0.49\textwidth}
    \setlength{\tabcolsep}{0.25em}
    \centering
    \begin{tabular}{m{\mywidth} m{\mywidth}}
        \includegraphics[trim={0.22cm 0.3cm 0.2cm 0.2cm},clip,width=\mywidth]{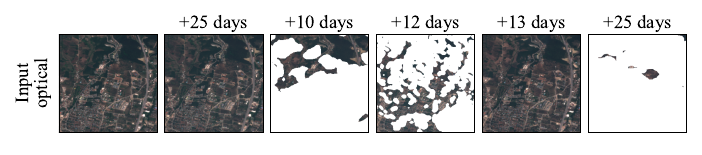} & \includegraphics[trim={0.22cm 0.3cm 0.2cm 0.2cm},clip,width=\mywidth]{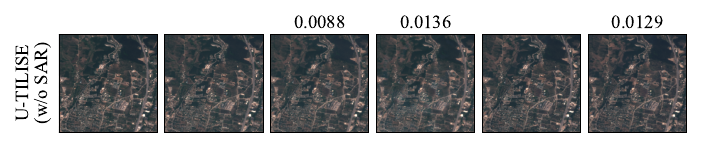}
        \\
        \includegraphics[trim={0.22cm 0.3cm 0.2cm 0.2cm},clip,width=\mywidth]{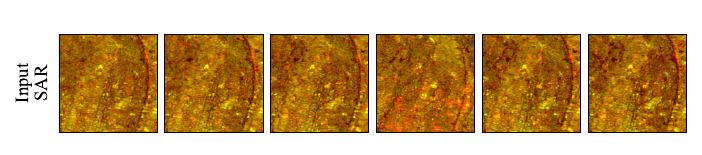} & \includegraphics[trim={0.22cm 0.3cm 0.2cm 0.2cm},clip,width=\mywidth]{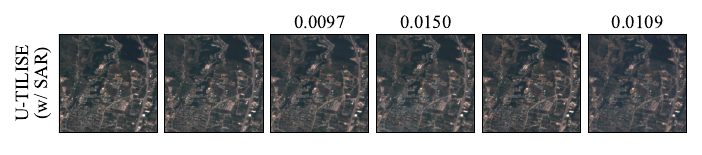}
        \\
    \end{tabular}
    \caption{Visual comparison of two \utilise variants on the \senXIImscrts dataset. On the left are the inputs for an exemplary sequence. For multi-spectral data, the RGB channels are displayed; SAR data are rendered as \mbox{two-channel} image composites (VV/VH amplitude). Numbers indicate the temporal spacing from the preceding image. \utilise prediction from only optical time series (top) and from combined optical and SAR input (bottom) are visually indistinguishable. Numbers are mean absolute errors over all masked pixels, across all 13~spectral bands.}
    \label{fig:sar}
\end{figure*}
We provide quantitative results in Table~\ref{tab:results_sar} and a visual example in Fig.~\ref{fig:sar}. As already observed with \earthnet, linear interpolation between the most recent and the next available observation performs significantly better than the more widely used baselines that replicate either the \emph{last} or the temporally \emph{closest} observation. 
\utilise achieves marginally better error metrics with SAR guidance than without, but the differences ($<\,$0.05\% of the intensity range in MAE and RMSE, $<\,$0.2$^\circ$ in SAM, $<\,$0.2\% in SSIM) are negligible and well within the stochastic fluctuations of deep network training. Moreover, the linear interpolation baseline is on par with both variants, and all three results remain well below the fidelity achieved on \earthnet (Table~\ref{tab:results_main}), although the numbers are not directly comparable since \senXIImscrts includes 13~bands of which 9 have been \emph{up}sampled to 10$\,$m GSD, whereas \earthnet consists of 4~bands that were \emph{down}sampled to 20$\,$m GSD. Upon inspection, we find that \senXIImscrts contains many sequences where the land cover is static and largely homogeneous. The results we obtain on \senXIImscrts neither confirm nor rule out a possible benefit through SAR guidance. We believe that a larger dataset with more non-linear temporal variations will be needed to carry out a conclusive comparison.

\section*{Acknowledgment}
\noindent This research is based upon work supported in part by the Office of the Director of National Intelligence (ODNI), Intelligence Advanced Research Projects Activity (IARPA), via Contract \#2021-21040700001.  The views and conclusions contained herein are those of the authors and should not be interpreted as necessarily representing the official policies, either expressed or implied, of ODNI, IARPA, or the U.S. Government.  The U.S. Government is authorized to reproduce and distribute reprints for governmental purposes notwithstanding any copyright annotation therein.

\bibliographystyle{IEEEtran}
\bibliography{bib}

\begin{IEEEbiography}
[{\includegraphics[width=1.0in,height=1.25in,clip,keepaspectratio]{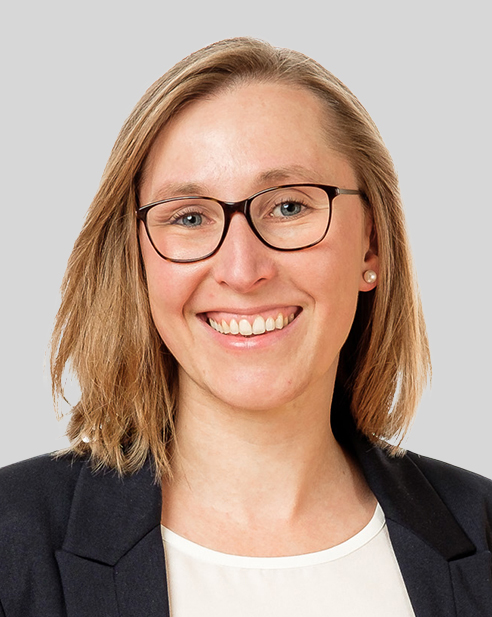}}]{Corinne Stucker}
received the B.Sc.\ degree in \mbox{geomatics} engineering and planning and the M.Sc.\ degree in geomatics engineering from ETH Z\"urich, Z\"urich, Switzerland, in 2015 and 2017, respectively, and the Ph.D.\ degree from ETH Z\"urich in July~2023, under the supervision of Prof.\ Dr.\ Konrad Schindler. Her research lies at the intersection of 3D~computer vision, machine learning, and remote sensing, focusing on scene representations for dense 3D~reconstruction from satellite imagery.\end{IEEEbiography}

\begin{IEEEbiography}
[{\includegraphics[width=1.0in,height=1.25in,clip,keepaspectratio]{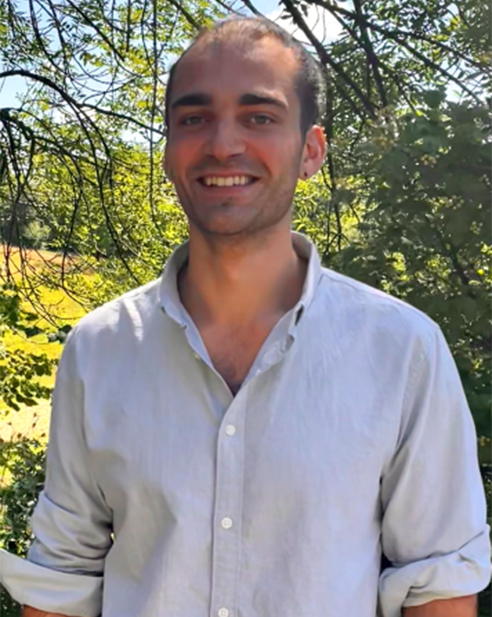}}]{Vivien Sainte Fare Garnot}
received the Ph.D.\ degree in geospatial computer vision from Gustave Eiffel University, Champs-sur-Marne, France (IGN, French mapping agency), in January~2022.

His thesis explored novel deep-learning architectures for crop type mapping from satellite image time series. Among other aspects, it focused on leveraging the temporal dimension of such data with attention mechanisms. The thesis also proposed a novel method to leverage the hierarchical structure of the crop type taxonomy and showed how to address crop type mapping as a panoptic segmentation task. He is currently a Post-Doctoral Researcher at the Institute for Computational Science, University of Z\"urich, Z\"urich, Switzerland, and also affiliated with the EcoVision Laboratory (since 2022). For his post-doctoral research, he is involved in the development of machine-learning methods for forest monitoring and other ecological applications. More broadly, he is interested in methods that leverage remote sensing data to improve our understanding of the Earth’s evolution.\end{IEEEbiography}

\begin{IEEEbiography}
[{\includegraphics[width=1.0in,height=1.25in,clip,keepaspectratio]{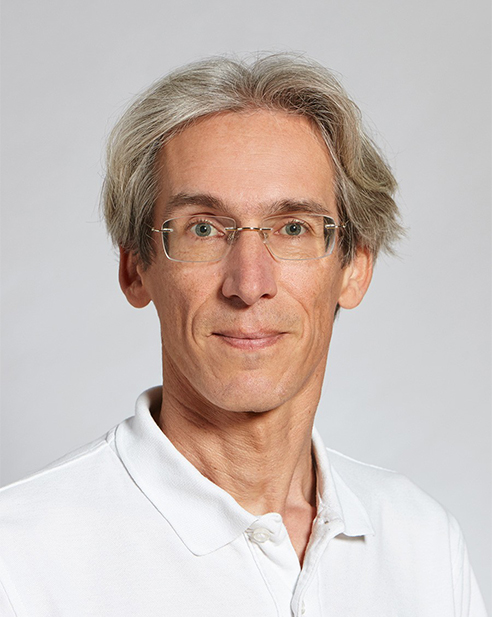}}]{Konrad Schindler}
(Senior Member, IEEE) received the Diplom-ingenieur (M.Tech.) degree from the Vienna University of Technology, Vienna, Austria, in 1999, and the Ph.D.\ degree from the Graz University of Technology, Graz, Austria, in 2003.

He was a Photogrammetric Engineer in private industry and held research positions at the Graz University of Technology; Monash University, \mbox{Melbourne}, VIC, Australia; and ETH Z\"urich, Z\"urich, Switzerland. He was an Assistant Professor of image understanding with TU Darmstadt, Darmstadt, \mbox{Germany}, in 2009. Since 2010, he has been a tenured Professor of \mbox{photogrammetry} and remote sensing with ETH Z\"urich. His research interests include remote sensing, photogrammetry, computer vision, and machine learning.\end{IEEEbiography}

\end{document}